\documentclass[Afour,sageh,times]{sagej}

\usepackage{amsmath,amsfonts}
\usepackage{amssymb}
\usepackage{array}
\usepackage[caption=false,font=normalsize,labelfont=sf,textfont=sf]{subfig}
\usepackage{textcomp}
\usepackage{stfloats}
\usepackage{url}
\usepackage{verbatim}
\usepackage{graphicx}
\usepackage[table]{xcolor}
\usepackage{xcolor}
\usepackage{multirow}
\usepackage{bm}
\usepackage{mathtools}
\usepackage{makecell}
\usepackage{adjustbox}
\usepackage{pifont}
\usepackage{lastpage}

\newenvironment{loa}[1]
{\subsection*{\normalsize\sagesf\bfseries List of Abbreviations}\begin{refsize}\noindent #1}
{\end{refsize}} 

\usepackage{datatool}
\newcommand{\sortitem}[1]{
  \DTLnewrow{list}
  \DTLnewdbentry{list}{description}{#1}
}
\newenvironment{sortedlist}{
  \DTLifdbexists{list}{\DTLcleardb{list}}{\DTLnewdb{list}}
}{
  \DTLsort{description}{list}
  \begin{itemize}
    \DTLforeach*{list}{\theDesc=description}{
      \item[] \theDesc}
  \end{itemize}
}

\usepackage{scalerel}

\usepackage[T1]{fontenc}
\UseRawInputEncoding

\usepackage[noend]{algpseudocode}
\usepackage{algorithmicx,algorithm}
\usepackage[normalem]{ulem}
\usepackage[switch,  mathlines]{lineno}
\usepackage{comment}
\usepackage[colorlinks,bookmarksopen,bookmarksnumbered,citecolor=red,urlcolor=red]{hyperref}

\def\volumeyear{2023}

\begin{document}

\runninghead{Tassi \textit{et al.}}

\title{IMA-Catcher: An IMpact-Aware Nonprehensile Catching Framework based on Combined Optimization and Learning
\vspace{-.5cm}
}

\author{Francesco Tassi\affilnum{1}\affilnum{+}, Jianzhuang Zhao\affilnum{1}\affilnum{+}, Gustavo J. G. Lahr\affilnum{1},  Luna Gava\affilnum{2}, Marco Monforte\affilnum{2}, \\ Arren Glover\affilnum{2}, Chiara Bartolozzi\affilnum{2}, and Arash Ajoudani\affilnum{1}
\vspace{-1cm}
}

\affiliation{\affilnum{1}{\scriptsize Human-Robot Interfaces and Interaction Lab., Istituto Italiano di Tecnologia, Italy}\\
\affilnum{2}{\scriptsize Event-Driven Perception for Robotics Lab, Istituto Italiano di Tecnologia, Italy}
\affilnum{+}{\scriptsize Contributed equally to this work.}
}

\corrauth{Francesco Tassi, Istituto Italiano di Tecnologia, francesco.tassi@iit.it}

\begin{abstract}
Robotic catching of flying objects typically generates high impact forces that might lead
to task failure and potential hardware damages.
This is accentuated when the object mass to robot payload ratio increases, given the strong inertial components characterizing this task.
This paper aims to address this problem by proposing an implicitly impact-aware framework that accomplishes the catching task in both pre- and post-catching phases. In the first phase, a motion planner generates optimal trajectories that minimize catching forces, while in the second, the object's energy is dissipated smoothly, minimizing bouncing.
In particular, in the pre-catching phase, a real-time optimal planner is responsible for generating trajectories of the end-effector that minimize the velocity difference between the robot and the object to reduce impact forces during catching.
In the post-catching phase, the robot's position, velocity, and stiffness trajectories are generated based on human demonstrations when catching a series of free-falling objects with unknown masses.
A hierarchical quadratic programming-based controller is used to enforce the robot's constraints (i.e., joint and torque limits) and create a stack of tasks that minimizes the reflected mass at the end-effector as a secondary objective.
The initial experiments isolate the problem along one dimension to accurately study the effects of each contribution on the metrics proposed. We show how the same task, without velocity matching, would be infeasible due to excessive joint torques resulting from the impact. The addition of reflected mass minimization is then investigated, and the catching height is increased to evaluate the method's robustness. Finally, the setup is extended to catching along multiple Cartesian axes, to prove its generalization in space.
\vspace{-.5cm}
\end{abstract}

\keywords{Nonprehensile robotic catching, impact awareness, optimal planning and control, learning from human demonstrations.}

\maketitle
\section{Introduction} \label{sec:intro}

Dynamic robotic manipulation tasks are challenging since they require tight coordination between object detection, motion planning, and control. Activities such as throwing~\citep{tossingbot2020tro}, hitting~\citep{tennis2020iros}, juggling~\citep{juggling2021iros}, and catching~\citep{kim2014catching,softcat2016tro} are only some of the applications that characterize multiple fields, ranging from logistics to aerospace. Catching flying objects is particularly challenging, as the overall catching motion demands fast and precise planning with quick execution. 
The catching task presents an additional and so far overlooked challenge of reducing the impact force between the robot and the object during the contact phase, as it can cause physical damage to both sides.

Catching of flying objects can be classified into prehensile and nonprehensile, based on how the object is intercepted by the robot. In the former case, the closure's force is exerted as bilateral constraints, clamping the object.  
In the latter, it is not possible to prevent any infinitesimal motion of the object nor to resist to any of the applied external wrenches, hence it is defined as a task only subject to unilateral constraints \citep{Siciliano_nonprehensile_review}. Hence a nonprehensile catching task can be defined as the scenario in which the grasp of the object is not restrained by neither \textit{form closure} nor \textit{force closure} \citep{prattichizzo2016grasping}.
For example, when using a stiff robot, the object may bounce away and lose contact, resulting in a second and possibly multiple impacts in the post-impact phase ~\citep{ajoudani2012tele}, 
thus posing additional challenges \citep{Siciliano_nonprehensile_review}.

\begin{figure}[H]
    \centering
    \includegraphics[width=\linewidth]{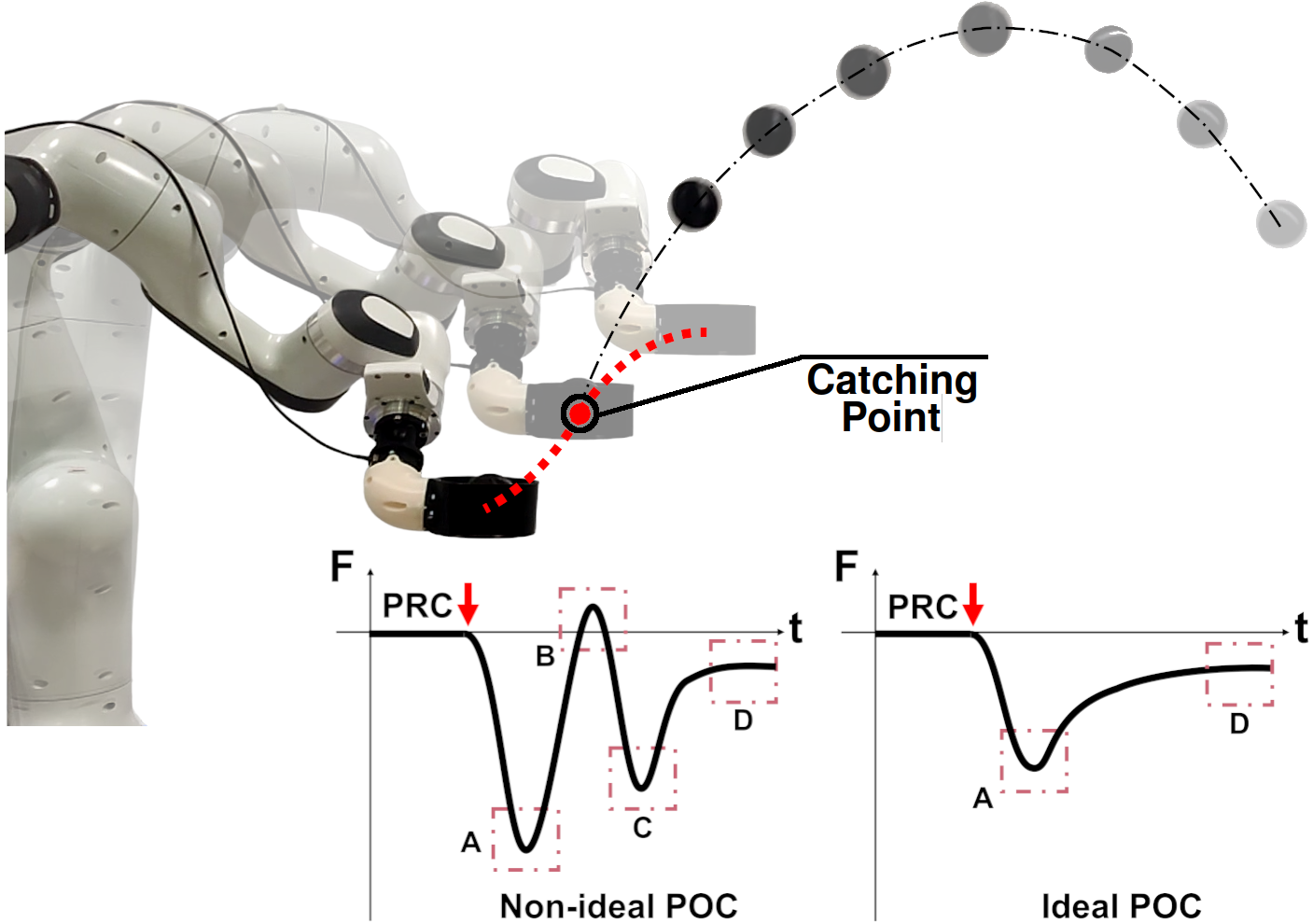}
    \caption{An ideal nonprehensile stiff catch is defined by a low impact force peak in PRC and no bouncing in POC (bottom right). In non-ideal catching (bottom left), after the impact (A) the object bounces and loses contact with the end-effector (B), resulting in a second (C) or multiple impact peaks before reaching steady state (D). IMA-Catcher aims to reduce the impact force (A) via velocity matching in PRC and reduce objects' bouncing, keeping stable contact in POC until steady state.
    }
    \label{fig:2Doverlay}
\end{figure}

Inspired by the works on reference spreading \citep{reference_spreading}, the nonprehensile catching motion can be divided into two phases: Pre-Catching (PRC) and Post-Catching (POC). 
PRC is the period preceding the first contact between
the object and the robot. When the object and the End-Effector (EE) collide, an impact is generated, the PRC phase ends and the POC starts. This is defined as the period following the impact, until steady state (see Fig.\ref{fig:2Doverlay}). 
Even though PRC and POC are isolated from a modeling perspective, the success of POC strongly relies on the performance of PRC since the impact force mainly depends on the relative robot-object velocity and their reflected inertia \citep{wangmodel22, Walker1994}.

PRC's main objective is to minimize the impact force by reducing the relative velocity between the robot and the object using velocity matching (VM). However, fast real-time object detection, trajectory estimation, and a computationally-efficient online planning algorithm are preconditions for the PRC, as the period of PRC is extremely short. Due to the high velocity of the flying object, a fast response motion of the robot is required to achieve VM while still satisfying both the robot's kinematic and dynamic limitations (i.e., maximum velocities, accelerations, torques).
To this purpose, an online planner during PRC is necessary, where the planned interception point is updated in real-time based on the estimated object's trajectory, which becomes more accurate as time progresses.

Meanwhile, in the POC phase the energy injected by the impact should be dissipated without overloading the robot's internal torques and avoiding bouncing as far as possible. It is important to note that the stiffness parameters of the robot's impedance controller - commonly used in interaction tasks - do not affect the impact force due to the extremely short impact period, i.e., a few milliseconds \citep{wangmodel22, haddadin2009requirements}. Still, these are fundamental to dissipate the post-impact forces~\citep{haddadin2009requirements,ajoudani2012tele}, follow a desired post-catching trajectory \citep{ajoudani2012tele} and prevent damaging the object \citep{Uchiyama2012egg} or the robot, especially in tasks with high speed \citep{Tassi_Gholami2021} and/or high object's mass \citep{stouraitis2020multiHybrid}. To dissipate the post-impact forces in POC, a compliant behavior was proven helpful~\citep{haddadin2009requirements} but challenging~\citep{abu2020variable, Zhao2022}. As humans can catch different flying objects safely and naturally by actively regulating the upper arm stiffness, even without knowing the exact weight of the objects~\citep{ajoudani2012tele}, we leverage learning from human demonstrations (LfD) and transferring the human behavior to the robot for the POC phase.

Our main focus is to show that, even in catching applications, successfully catching might not be enough and many other aspects should be considered simultaneously i.e., VM. Another aspect, for instance, is the Reflected Mass Minimization (RMM) at the EE, which we employ as an added layer to implicitly minimize the contact forces generated by the impact. Indeed, by exploiting the redundancy of the robot, and by estimating the impact's direction, it is possible to reconfigure the robot's redundancy to implicitly optimize, from a kinematics perspective, the robot's capacity to withstand impacts~\citep{Walker1994, tassi2022} by optimally transmitting the reaction forces to the ground while minimizing the generation of internal torques.
In addition, we evaluate the RMM's role in both PRC and POC stages, since smaller reflected inertia leads to a more uniform internal torque distribution, avoiding torque peaks that might lock and even damage the robot~\citep{Haddadin2012ijrr}.

\subsection*{Contributions}

To the best of our knowledge, this is the first time that a nonprehensile catching task (Fig.~\ref{fig:2Doverlay}) has been considered from an impact-aware point of view. While prehensile applications are limited to catching, nonprehensile scenarios are wider and include: throwing \citep{pizzatoss, tossingbot2020tro}, batting/redirecting falling objects on the fly (e.g., sports \citep{senoo2006ball, tennis2020iros}, fast objects sorting/separation), bimanual manipulation (e.g., juggling \citep{9341775}, dynamic hand-to-hand objects' handover, grabbing and tossing \citep{BOMBILE2023104481}), dynamically pushing objects \citep{stouraitis2020multiHybrid, 5509491}, sliding/rolling objects on the EE's surface for manipulation \citep{4543574, juggling2021iros, taylor2019optimal, 5979782}. Recent studies even investigate the similarities between bipedal locomotion and nonprehensile manipulation \citep{FARID202251}.

In this work, we study a nonprehensile catching task, specifically focusing on a thorough analysis of the exchanged forces, which is not treated in the literature \citep{basketball_catch, 7139529, juggling2021iros} but can easily lead to task failure and hardware damages, due to the high peak torques generated by the catching impact at the EE.
Hence, we decide to shift our focus from grasping- to impact-related challenges and issues.
To this purpose, we opt for a basket-like EE and a spherical object,
to facilitate the repeatability of impact forces and the consistency between trials (avoiding random misalignments between asymmetric objects and the EE). We do not employ a purely planar EE, to avoid the issue of stabilizing the ball on a flat surface, which is a topic that has been plentifully investigated in the literature \citep{ballbeam, 4543574, juggling2021iros, 5979782} and that is not part of our goal. Indeed, we are mostly interested on how aspects such as VM, impedance regulation and RMM can affect the catching impact.
Indeed, in most catching applications, it is necessary to deal with high-impact forces, which can affect task success and potentially damage the robot.
We employ a robot with small maximum torques compared to the ones generated during such a dynamic task and show that the proposed strategy can overcome these limitations.

In our previous work~\cite{zhao2022impact}, we started addressing the impact-aware nonprehensile catching problem.
In the PRC phase, however, we solved a different VM problem offline. This Quadratic Programming (QP) problem minimized the difference between the entire object's (estimated) and robot's trajectories without hard constraints on the final catching point. Besides, we tested only a single-axis scenario (vertical dropping) using a Cartesian space impedance controller, which did not ensure general applicability to multi-axis scenarios.

With the IMA-Catcher instead, we are able to extend the catching task to a multi-axis scenario, thanks to a different formulation of the VM optimal trajectory generation problem and to its real-time feature that exploits the high-frequency data of the event-driven camera~\citep{Posch_etal2011}.
Specifically, we treat the impact in an implicit way, where the impact model is not directly formulated in the overall proposed framework.
Rather, we exploit the one-dimensional frictionless model to formulate a hierarchical QP-based metric that enables RMM.
Besides, we are not explicitly modeling the state jumps induced at joint velocity and torque levels, since we rely on the local stability of the lower-level joint impedance controller \citep{opensot} and on the maximum external perturbation being limited by the small object's mass.

Merging multiple elements under an overarching framework, we individually address some currently open challenges, such as: the solution of both problems of i) optimal planning and ii) VM, in a real-time efficient manner, while focusing on the impact forces generated. Indeed, most approaches are either purely focused on the kinematics and tracking accuracy aspect of the catching \citep{softcat2016tro} without considering impact forces and RMM or employ complex nonlinear optimization techniques~\citep{Bauml2010hirzinger} which does not enable real-time feasibility.
The main advantages with respect to existing studies lie in the lightweight computational burden of the proposed optimal planner, which is a critical issue for the  success of a catching task, together with the addition of a human-inspired POC phase, to dissipate impact energy and avoid bouncing issues.
Most studies only tackle these problems separately, neglecting important aspects of the catching task, which should be considered via a holistic approach.
A detailed comparison with previous literature can be found in Table~\ref{tab:lite_compare} and is further articulated in Sec.~\ref{sec:related-works}.

As a control strategy, we exploit a Hierarchical Quadratic Programming (HQP) controller to minimize the reflected mass at the EE as a secondary strict task and compare the framework's performance by evaluating the experiments in free-falling (single-axis) and multi-axis scenarios.

The main contributions of this paper are summarized as follows.
\begin{enumerate}
    \item An autonomous robotic catching framework that seamlessly unifies strategies for smoothly handling objects before and after the catch, providing an end-to-end solution for robotic object manipulation;
    \item PRC planning: a lightweight QP-based real-time planner for VM, to minimize impact forces while enforcing kinematic constraints; 
    \item POC planning: online mapping and scaling of the learned single-axis human stiffness to multi-axis, to dissipate contact energy, avoid bouncing and ensure a successful catch of the object;
    \item Multi-axis investigation on the effect of hierarchical RMM, stiffness modulation and VM, on the realization of high force-to-payload ratio impacts, otherwise unfeasible due to robot's low torque limits. Robustness analysis against relative catching distance, perceived mass at the EE and dimensionality extension.
\end{enumerate}

It is important to re-emphasize that in this work we first deliberately isolate the problem along one dimension to accurately study the effects of the impact on the dynamic interaction metrics during and after the impact, given our design choices. Eventually, we extend the proposed planning and control strategies to multi-axis scenarios, as demonstrated by the two-dimensional setup. A three-dimensional case is not considered due to the technological limitations related to the vision system employed.

The reader should note that, the focus of this study, is to perform an in-depth evaluation on how certain factors (i.e., Cartesian stiffness at the EE, optimal VM, RMM) can affect catching-induced impacts, and how this translates to the torque peaks generated on the motors, eventually mitigating/optimizing them by minimizing the reflected mass perceived at the EE.
With this view in mind, all the assumptions (i.e., one-dimensional impact model, horizontal EE orientation) and design choices (basket-like EE and spherical object) that will be further discussed in the remainder, have been taken to best isolate each aspect under study, given the multiple factors that characterize such a complex task, that would potentially affect the experimental process. Therefore, we strongly believe that this work can only be considered as a piece of a more elaborate all-encompassing framework that deserves further studies, eventually capable of practically achieving optimal catching while addressing all the critical aspects of such a complex task.

The overall article is organized as follows. The background and related works are discussed in Sec. \ref{sec:related-works} to highlight the novelty of the proposed method. The overall methodology is presented in Sec. \ref{sec:methodology}. The experiments, results, and discussions are presented in Sec.s \ref{sec:experimental-setup}-\ref{sec:experiments}, whereas Sec. \ref{sec:con} concludes this paper.
\section{Background and Related work}\label{sec:related-works}

This section introduces the nominal rigid-body impact model to analyze catching forces exchange. The research on flying object detection and trajectory estimation is then discussed, followed by the state of the art on flying objects-catching, divided into pre- and post-catching oriented studies. In addition, to address the novelty of the proposed method, Table \ref{tab:lite_compare} compares our method with the most relevant works in the literature, where the differences are discussed in more depth in the next sections.

\subsection{Impact Model} \label{sec:impact-model}

In this section, the robot dynamics is described to analyze the variables that affect the impact forces. Hence, the impact model is derived, which will be used to minimize the reflected mass at the EE via the hierarchical controller in Sec.~\ref{sec:mrm}.

The robotic manipulator rigid body model in joint space is
\begin{equation} \label{eq:rigid_body_full_eq}
    \boldsymbol{M}(\boldsymbol{q}) \ddot{\boldsymbol{q}} + \boldsymbol{C}(\boldsymbol{q},\dot{\boldsymbol{q}}) + \boldsymbol{g}(\boldsymbol{q}) = \boldsymbol{\tau} + \boldsymbol{\tau}_{ext},
\end{equation}
where $\boldsymbol{q}, \dot{\boldsymbol{q}}, \ddot{\boldsymbol{q}} \in \mathbb{R}^n $ are the position, velocity, and acceleration vectors in joint space, respectively, and $n$ is the number of degrees of freedom (DoF). $\boldsymbol{M}(\boldsymbol{q}) \in \mathbb{R}^{n \times n}$ is the joint space inertia matrix, $\boldsymbol{C}(\boldsymbol{q}, \dot{\boldsymbol{q}}) \in \mathbb{R}^{n}$ the Coriolis term, $\boldsymbol{g}(\boldsymbol{q})\in \mathbb{R}^{n}$ the gravity term, and $\boldsymbol{\tau} \in \mathbb{R}^{n}$ are the actuation torques. The external torques generated by the interaction forces with the environment are given by $\boldsymbol{\tau}_{ext} = \boldsymbol{J}^T(\boldsymbol{q}) \boldsymbol{F}_{ext}$, being $\boldsymbol{J}^T(\boldsymbol{q}) \in \mathbb{R}^{n \times 6}$ the transpose of the contact Jacobian matrix.
The model \eqref{eq:rigid_body_full_eq} may be rewritten in Cartesian space to facilitate the controller design 
\begin{equation}\label{eq:rigid_body_cartesian}
    \boldsymbol{\Lambda}(\boldsymbol{x}) \ddot{\boldsymbol{x}} + \boldsymbol{\mu}(\boldsymbol{x}, \dot{\boldsymbol{x}}) \dot{\boldsymbol{x}} + \boldsymbol{F}_g(\boldsymbol{x}) = \boldsymbol{F}_{\tau} + \boldsymbol{F}_{ext}.
\end{equation}
where $\boldsymbol{x}, \dot{\boldsymbol{x}},\ddot{\boldsymbol{x}} \in \mathbb{R}^{6}$ are the robot's pose, velocity and acceleration, respectively, $\boldsymbol{\Lambda} = \boldsymbol{J}^{-T}\boldsymbol{M}\boldsymbol{J}^{-1}$ is the inertia matrix, $\boldsymbol{\mu}(\boldsymbol{x}, \dot{\boldsymbol{x}}) = \boldsymbol{J}^{-T}(\boldsymbol{C}-\boldsymbol{M}\boldsymbol{J}^{-1}\dot{\boldsymbol{J}})\boldsymbol{J}^{-1}$ is the Coriolis term, $\boldsymbol{F}_g(\boldsymbol{x})=\boldsymbol{J}^{-T}(\boldsymbol{q})\boldsymbol{g}(\boldsymbol{q})$ the gravity, and $\boldsymbol{F}_{\tau} = \boldsymbol{J}^{-T}(\boldsymbol{q}) \boldsymbol{\tau}$ the actuation forces.

During the PRC, before the contact instant $t_c$, the robot's velocity is $\dot{\boldsymbol{x}}$ and the object's velocity is $\dot{\boldsymbol{x}}_{o}\in \mathbb{R}^{6}$. After the impact, the robot's and the object's velocities update to $\dot{\boldsymbol{x}} + \delta\dot{\boldsymbol{x}}$ and $\dot{\boldsymbol{x}}_{o} + \delta\dot{\boldsymbol{x}}_{o}$, respectively.
Given the short time frame of the impact phase, the control actions cannot produce an immediate effect, therefore, the robot's velocity and configuration will mostly dictate the reactive behavior.
For an instantaneous collision, given that the direction of contact is defined by the normal vector $\bm{u} \in\mathbb{R}^3$, the coupled dynamics of the robot and the object as rigid bodies are described by:
\begin{equation}
    [(\dot{\boldsymbol{x}} + \delta\dot{\boldsymbol{x}})-(\dot{\boldsymbol{x}}_{o} + \delta\dot{\boldsymbol{x}}_{o})]^T \boldsymbol{u} = -e(\dot{\boldsymbol{x}}-\dot{\boldsymbol{x}}_{o})^T \boldsymbol{u},
    \label{eq:restitution_coefficient}
\end{equation}
with $0<e<1$ being the coefficient of restitution. When $e=1$ is said to be an elastic collision, the bodies have maximum velocity after the impact; while $e=0$ is a plastic collision, i.e., the relative velocity of the two bodies is zero. 

The interaction forces generate finite impulsive forces $\hat{\boldsymbol{F}}$ at the contact point: $\hat{\boldsymbol{F}}=\lim_{\delta t \rightarrow 0} \int^{t+\delta t}_{t} \boldsymbol{F}_{ext}(s) ds$, where $\delta t$ is the duration of the impact.
From the integration of (\ref{eq:rigid_body_full_eq}), the analysis of impact shows that the variation of velocity in the robotic manipulator due to the impact is given by $\delta \dot{\boldsymbol{x}} = \boldsymbol{\Lambda}^{-1} \hat{\boldsymbol{F}}$ \citep{impactAnalysis2000kim, impactAnalysis1990walker}.
Assuming the mass of the object is ${m}_{o}$, its velocity variation due to impact is $\delta \dot{\boldsymbol{x}}_{o}= 1/m_{o} (-\hat{\boldsymbol{F}})$ \citep{advancedDynamics2003Greenwood}.
Finally, substituting both relations into (\ref{eq:restitution_coefficient}), the one-dimensional frictionless contact impulse model becomes:

\begin{equation}
    \hat{F} =\frac{-(1+e)(\dot{\boldsymbol{x}}-\dot{\boldsymbol{x}}_{o})^T\bm{u}}{\bm{u}^T(\boldsymbol{\Lambda}^{-1} + \frac{1}{m_{o}} \boldsymbol{I})\bm{u}}.
    \label{eq:impact_relationship}
\end{equation}

Equation (\ref{eq:impact_relationship}) shows the dependencies that can minimize the exchanged force at the moment of impact. First, it is visible that the VM is predominant in minimizing the forces as the relative velocity between the robot and the object are proportional to the impact force model. 
Being the robot's EE velocity bounded, it is hard to deal with falling objects, as they will quickly reach velocities higher than the maximum speed of the robot's EE.

The second term that affects impact forces is the object's mass and the robot's reflected inertia. Although we cannot control the mass of the flying object, the robot's inertia matrix $\bm{\Lambda}$ can be modulated. 
According to the results in~\cite{wangmodel22}, for pure-torque controlled robots or under-actuated pendulums, the reflected robot's mass along the impact direction can be expressed as: 
\begin{equation} \label{eq:rmm_def}
    m_{u}(\bm{q}) = \begin{bmatrix} \bm{u}^T \bm{\Lambda}_v^{-1} \bm{u} \end{bmatrix}^{-1}
\end{equation}
where $\bm{\Lambda}_v\in\mathbb{R}^{3 \times 3}$ is the positional part of the decomposition of the inertia matrix $\bm{\Lambda}$. 
In \cite{Haddadin2012ijrr}, a distinction between blunt and sharp contacts is studied, and the minimization of the reflected mass plays a relevant role in the interaction forces.
The dependency of the inertia matrix from the robot's configuration via $\bm{J}$ can therefore be exploited as in \cite{Walker1994, tassi2022} to achieve RMM as it will be explained in Sec. \ref{sec:mrm}, where~\eqref{eq:rmm_def} is used to construct a secondary hierarchical layer, which instantaneously minimizes the reflected mass at the EE, as if the impact should occur at the next time instant. This provides greater robustness against contact timing.

\begin{table}[!t]\caption{Comparison with relevant prior works.\\
$^*$Referred to the 'direct catching' method in the paper.}
\setlength\extrarowheight{3.5pt}
\begin{center}\label{tab:lite_compare}

\resizebox{\columnwidth}{!}{\begin{tabular}{c|ccc|cc|c}
\hline
\multirow{2}{*}{Paper\textbackslash Feature} & \multicolumn{3}{c|}{PRC} & \multicolumn{2}{c|}{POC}  & \multirow{2}{*}{RMM}\\[0.7mm]
        \cline{2-6}
        & \multicolumn{1}{c|}{VM} & \multicolumn{1}{c|}{Optimal} & \multicolumn{1}{c|}{Real-time} & \multicolumn{1}{c}{Online} &  \multicolumn{1}{|c|}{VIC} &  \\
\hline
\citet{basketball_catch}$^*$  & \ding{51} & \ding{53} & \ding{51}  & \ding{51} & \ding{53}  & \ding{53}\\[0.7mm]
\citet{Bauml2010hirzinger}  & \ding{53} & \ding{51} & \ding{51}  & \ding{53} & \ding{53}  & \ding{53}\\[0.7mm]
\citet{kim2014catching}  & \ding{53} & \ding{53} & \ding{51}  & \ding{53} & \ding{53}  & \ding{53}\\[0.7mm]
\citet{softcat2016tro} & \ding{51} & \ding{51} & \ding{51}  & \ding{53} & \ding{53} & \ding{53}\\[0.7mm]
\citet{poly2015brun}& \ding{53} & \ding{53} & \ding{51}  & \ding{53} & \ding{53}  & \ding{53}\\[0.7mm]
\citet{sato2020high}& \ding{53} & \ding{53} & \ding{53}  & \ding{53} & \ding{53}  & \ding{53}\\[0.7mm]
\citet{wang2022ev} & \ding{53} & \ding{53} & \ding{51}  & \ding{53} & \ding{53} & \ding{53}\\[0.7mm]
\textbf{Ours} & \ding{51} & \ding{51} & \ding{51}  & \ding{51} & \ding{51} & \ding{51}\\[0.7mm] 
\hline
\end{tabular}}
\end{center}
\vspace{-2mm}
\end{table}

\subsection{Pre-catching}
The PRC phase presents two main issues: i) finding a catching point and ii) planning a trajectory to it. For prior works that ignore the impact effect during the contact phase and do not consider the POC \citep{kim2014catching, Andersson1989, senoo2006ball, poly2015brun, sato2020high, Bauml2010hirzinger, opball2011jan}, the catching point can be chosen along the estimated object's trajectory and inside the feasible task space of the robot. For example, in \cite{basketball_catch,poly2015brun}, the catching point is selected by considering robot's joint limits and manipulability; while in \cite{nonprehensile_caging} a simple catching policy is defined, which however is non-optimal and does not consider impacts. Instead, VM is the main concern when choosing the catching point for the works that consider the impact \citep{basketball_catch, zhao2022impact}. 
Differently, the optimality of the trajectories in the PRC phase cannot be ensured since there is no explicit optimization formulation used in \cite{basketball_catch}. In our previous work \cite{zhao2022impact}, we identify the optimal catching point based on the estimated trajectory of the object through a QP formulation.

Once the catching point is identified, a smooth trajectory must be planned from the initial point to the desired catching point. The model-based polynomial method~\citep{Andersson1989, senoo2006ball, poly2015brun, sato2020high} can give a quick solution because of its computational efficiency. However, this method cannot ensure the robot's trajectory is feasible since the constraints of the robot in Cartesian and joint space are not directly considered. The optimization-based approach can address this issue via kinematic constraints instead~\citep{opball2011jan}.
These time-driven methods are particularly sensitive to inaccurate estimation of flying objects. Therefore, real-time re-planning can relieve this problem by updating the object's trajectory using online perception and increasing robustness.

Other solutions for PRC take inspiration from humans, which can catch moving objects on the fly with non-zero velocity, generating fast yet smooth trajectories, since LfD approaches produce fast and reactive motions~\citep{bennett1994insights,kajikawa1999analysis}. In \cite{kim2014catching}, the robot motion is learned from throwing demonstrations, but the robot stops immediately upon object contact with the EE, which may result in the bouncing of the object and high force peaks. To address this issue, in \cite{softcat2016tro}, the robot continues to track the predicted objects' path after contact, generating a soft-catching motion. However, the actual state (i.e., position, velocity) of the robot and the flying object are not presented, making it difficult to evaluate the performance of VM. Furthermore, a multi-finger gripper is used to catch the flying objects, where the form closure is generated once the objects are caught.
Hence, LfD studies have generated promising results, yet they did not consider the post-impact vibrations during POC, limiting the catching to lightweight objects.

We propose a novel QP formulation to plan an optimal trajectory for VM, subject to the robot's constraints, to catch the flying object successfully. This is done in real-time, based on the object's estimated trajectory, increasing robustness to the uncertainties in the state estimates.

\subsection{Post-catching}

In the ideal case where the restitution coefficient $e$ is zero (the object would completely stick to the EE), the POC strategy would reduce to simply stopping the robot's motors a time instant after the contact occurs, i.e., achieving good VM and RMM during PRC would be enough, which was implemented in ~\cite{basketball_catch}.
In the practical case ($e>0$) the impact will dissipate energy, and the object will bounce, each time with a velocity closer to the EE's velocity. Hence, compliant behavior in POC is paramount for dissipating energy and achieving a successful task. Compliance is also important in dealing with uncertainties, either related to the object's mass or inaccuracies in the perception or tracking, where a stiff controller would lead to higher interaction forces. Compliance finally improves stability, since bouncing introduces discontinuities in the controller, which could lead to unstable behavior.

It is possible to obtain highly damped behaviors using impedance \citep{Senoo2016plasticImpedance} or admittance \citep{Fu2021admittance} controllers, but so far, only fixed-parameters controllers have been proposed with empirically chosen values. In \cite{stouraitis2020multiHybrid}, a hybrid controller was implemented with multi-mode trajectory optimization to halt a heavy object traveling on a table at a constant speed of $0.88m/s$. However, due to the transition from free to coupled motion, the switching between different modes results in large nonlinearities, with a heavy computational burden. Besides, initializing the whole system is nontrivial. Hence, it is challenging to apply this method for catching flying objects with high velocity, requiring a prompt response from the robot. A recent research~\cite{yantro2024}, proposed a catching system based on the improved version of the impact-aware multi-modal trajectory optimization approach in~\cite{stouraitis2020multiHybrid}, which optimized the stiffness and desired position for a low-level cartesian space indirect-force controller. Besides, the contact selection optimization module chooses the desired contact modality based on the flying object trajectory prediction and impact model. Based on this system, two KUKA-iiwa robots successfully caught moving/flying boxes. Although the presented results were impressive, these can only relate to prehensile catching, since the closure's force is exerted once the boxes are caught by the two robots.
Furthermore, due to the overall catching system not considering the robot joint limits and singularity configurations, the robots were likely to stop in the middle of the catching process, as discussed by the same authors.

In the real world, humans have the ability to efficiently respond to unstable dynamics by adaptive control of their impedance~\citep{burdet2001central}.
Therefore, similarly to VM in PRC, the control principles during the POC phase can also be inspired by humans. For example, the human-like force and impedance adaptation method proposed in~\cite{yang2011humanmotor} enables the robot to interact with unknown environments.
Tele-impedance was proposed in our previous work~\cite{ajoudani2012tele}, where EMGs estimate the human upper arm stiffness in real time by remotely controlling a follower robotic arm in a ball-catching experiment.
The results show that the Variable Impedance Controller (VIC) outperforms the constant control under several metrics.
To transfer the human VIC behavior to autonomous robots, LfD is applied.
For example, LfD setups have been used to teach impedance profiles in quasi-static tasks \citep{abu2018forceras, Wu2020aframe, Zhao2022}. Specifically for catching, the authors of \cite{phung2014learning} encoded the human kinematic motion with a Gaussian Mixture Model (GMM) and Gaussian Mixture Regression (GMR) to catch a flying ball. However, no impact forces were reported, and no information about human dynamics was encoded to improve the method. In conclusion, transferring humans' compliant behavior to autonomous robots for catching flying objects is still an open issue. 

In this work, we transfer the human variable stiffness profile to the robot by LfD, to absorb the post-impact forces, by properly scaling the behavior based on the differences between the two scenarios.
Please note that although robot and human kinematics differ, the trajectories and stiffness trends are estimated and replicated in Cartesian/operational space (considering the same weight-to-payload ratios) to avoid mapping inconsistencies. In addition, what is essential from human demonstrations is the stiffness profile behavior at the arm endpoint, to understand how it contributes to POC key performance indicators (impact forces, bouncing, etc.).
Indeed, we are interested in learning the overall shape of the stiffness profile, and how it varies across the different phases of a generic catching task.

\section{Methodology}\label{sec:methodology}

\begin{figure*}[t]
	\centering
	\includegraphics[trim=3.9cm 1.2cm 3.9cm 1.4cm,clip,width=\linewidth]{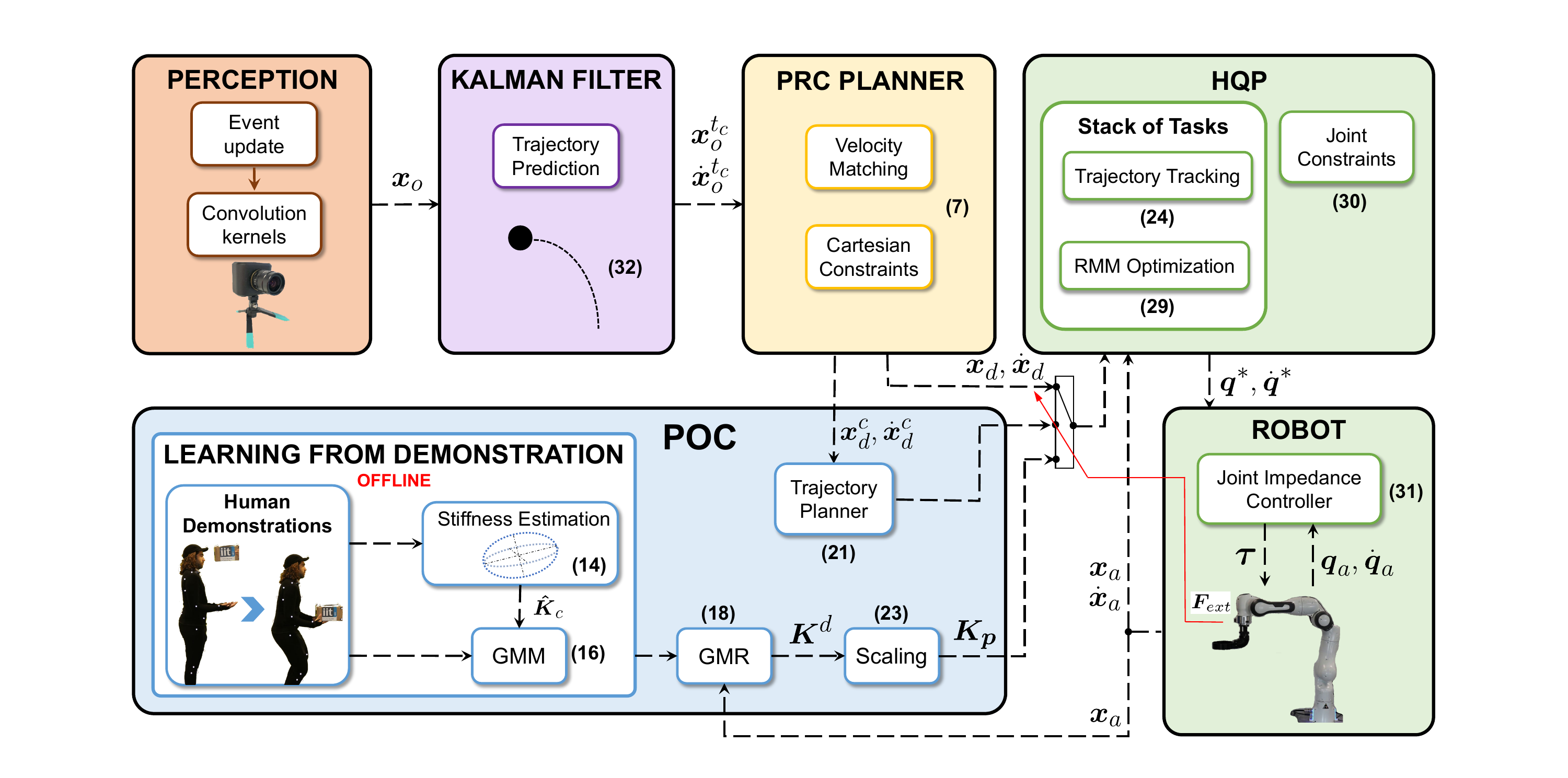}
	\caption{Summary diagram of the overall framework. From left to right, the object's trajectory identified in the perception module is sent to a Kalman filter to predict the catching instant. These are sent to the PRC optimization, which computes the PRC optimal catching trajectories via constrained QP for VM. In the POC phase, human demonstrations are used to learn the variable stiffness profile via GMM offline, and the trajectory is planned online with the scaled stiffness profile. Both the references generated in PRC and POC are provided to the HQP controller, which remains the same throughout. Lastly, the lower-level joint impedance controller generates the actuation torques provided to the robot. 
    }
	\label{fig:framework}
\end{figure*}

\subsection{Proposed framework}

We start by describing the overall structure of the proposed framework mentioned in contribution 1.
The complete scheme is reported in Fig.~\ref{fig:framework}, where, firstly, the perception module allows to detect the falling object via an event-camera, whose details in the implementation are described in Sec.~\ref{sec:Event-based perception}. The instantaneous position is provided to the trajectory estimation module, which uses a Kalman filter to estimate the entire trajectory and reduce noises, as explained in Sec.~\ref{sec:estimator}. Position and velocity estimated trajectories are sent to the PRC planner described in Sec.~\ref{sec:vm_planner}, where the proposed QP-based catching strategy is implemented in predictive fashion, generating the reference trajectories for the robot, that are passed to the HQP controller explained in Sec.~\ref{sec:method_hqp}. As soon as the catch is detected, the planning strategy smoothly switches to the POC module, which is instead described in Sec.~\ref{sec:poc} and allows to dissipate contact energy based on the learned human behavior. Its reference trajectories are sent to the same controller, which eventually provides the optimal joint trajectories to the lower level joint impedance controller, that generates the torque commands as in Sec.~\ref{sec:jic}.

\subsection{Pre-catching formulation}

\subsubsection{Optimal planner for velocity matching:} 

firstly, for the catch to be possible, it is necessary to assume that the object's trajectory does pass through the reachable workspace of the robot based on its initial pose and manufacturing constraints.
It is then necessary to outline the differences in terms of VM with respect to our previous formulation proposed in \cite{zhao2022impact}. 
The aim was to achieve optimal VM limited to the one-dimensional case with the QP problem:
\begin{equation} \label{eq:general_QP_formulation_old}
    \min_{\dot{\bm{x}}_R} \: \frac{1}{2} \left( \alpha \| 
    \dot{\bm{x}}_R-\dot{\bm{x}}_O \|^2 + \beta \| 
    \bm{x}_R-\bm{x}_O \|^2 - \gamma \| \bm{G} \bm{x}_R \|^2
    \right)
\end{equation}
where $\bm{x}_R,\dot{\bm{x}}_R \in\mathbb{R}^{6 T}$ are the actual position and velocity trajectories of the robot's EE written in vector form as: $
    {\bm{x}}_R = \begin{bmatrix}
        {\bm{x}}_{{i+1}}^T,
        {\bm{x}}_{{i+2}}^T,
        \hdots,
        {\bm{x}}_{{i+T}}^T
    \end{bmatrix}^T,
    \dot{\bm{x}}_R = \begin{bmatrix}
        \dot{\bm{x}}_{i}^T,
        \dot{\bm{x}}_{i+1}^T,
        \hdots,
        \dot{\bm{x}}_{i+T-1}^T
    \end{bmatrix}^T,
$
with $\bm{x}_{i}, \dot{\bm{x}}_{{i}} \in\mathbb{R}^{6}$ being the robot's Cartesian poses and velocities respectively at the $i$-th time instant and $T$ is the length of the prediction horizon, chosen based on the time necessary for the object to reach the ground. $\bm{x}_O,\dot{\bm{x}}_O \in\mathbb{R}^{6 T}$ are the trajectories of the object, which are structured analogously to those relative to the robot:
$
    {\bm{x}}_O = \begin{bmatrix}
        {\bm{x}}_{o_{i+1}}^T,
        {\bm{x}}_{o_{i+2}}^T
        ,..,
        {\bm{x}}_{o_{i+T}}^T
    \end{bmatrix}^T,
    \dot{\bm{x}}_O = \begin{bmatrix}
        \dot{\bm{x}}_{o_{i}}^T,
        \dot{\bm{x}}_{o_{i+1}}^T
        ,..,
        \dot{\bm{x}}_{o_{i+T-1}}^T
    \end{bmatrix}^T,
$
which are estimated based on the initial height of the free-falling object, with $\bm{x}_{o_{i}}, \dot{\bm{x}}_{o_{i}} \in\mathbb{R}^{6}$ being the object's poses and velocities respectively at the $i$-th time instant.
The cost function \eqref{eq:general_QP_formulation_old} establishes a soft hierarchical order via the weights $\alpha, \beta, \gamma \in\mathbb{R}$. The highest priority is to catch the object successfully (second cost in \eqref{eq:general_QP_formulation_old}), followed by VM ($\beta > \alpha$). The last term maximizes the catching height which inherently improves VM, since the object's falling velocity soon exceeds the maximum velocity of the robot's EE.
The selection matrix \resizebox{.85\columnwidth}{!}{$\bm{G} = diag \{ \left[0,0,1,0,0,0\right] ,\hdots, \left[ 0,0,1,0,0,0\right] \} \in\mathbb{R}^{6T \times 6T}$} isolates the $z-$components of $\bm{x}_R$ (reference frames in Fig. \ref{fig:1d-exp-setup}).

However, this formulation spans the entire period in which the object is in flight (requiring the entire object's trajectory), with no strict constraints on the catch (position and time) since the positional term is regulated by $\beta$, whereas the other objectives might compromise its success. Therefore, in predictive control fashion, \eqref{eq:general_QP_formulation_old} minimizes the overall differences (along the entire trajectories) in both position and velocity between the object and the robot's EE. This leads to the conflict between accurate catching position and VM, which is unsuitable for catching in three-dimensional space.

\subsubsection{Proposed formulation for scalable velocity matching:} \label{sec:vm_planner}

in this work we formulate an approach that is instead easily scalable to multi-axis catching scenarios.
We assume a fixed catching height at which the contact must occur, based on the following reasons.
Firstly, given the multiple aspects to analyze, related to both PRC and POC, it is important to fix as many external parameters as possible, to accurately isolate their individual effects on the metrics under study and to improve repeatability in terms of impact forces.
Indeed, catching in different points implies different arm configurations, therefore introducing variations in terms of joint torques. This affects the reflected mass at the EE \eqref{eq:rmm_def}, which would inhibit the analysis of the beneficial effects of VM and RMM on the catching forces.

Secondly, given the proposed POC strategy, the robot will require enough space to continue its movement after the catch. Therefore, the catching height is chosen as to avoid collisions with the table, while simultaneously maintaining high manipulability.

This assumption does not limit in any way the generality of the framework, since in applications where the height choice becomes important and must be optimized, it is possible to employ the first problem \eqref{eq:general_QP_formulation_old} to generate the optimal catching height based on the \textit{soft} priorities in its cost function, and eventually use this as input in the new planning problem \eqref{eq:general_QP_formulation}. For the sake of simplicity, we decided not to implement this in the present work, however this will be tackled in future developments.

As opposed to \eqref{eq:general_QP_formulation_old}, where the timespan $T$ refers to the object reaching the ground, in the proposed formulation the final time instant is the catching one $t_c$.
Under the assumption that the first-order Taylor approximation (Euler method) is valid for small planner's sampling period $\Delta t_p$, we can impose the catching position as a hard constraint through:
\begin{align}
    \min_{\dot{\bm{x}}_R} \: \frac{1}{2} \| 
    & \dot{\bm{x}}_r^{t_c} -\dot{\bm{x}}_o^{t_c} \|^2  \nonumber \\ 
    s.t. \hspace{.6cm} & \bm{x}_r^{t_c} = \bm{x}_o^{t_c} \nonumber \\
    & \bm{x}_R^{min} \le \bm{x}_{R_{0}} +\Delta t_p \ \dot{\bm{x}}_R \le \bm{x}_R^{max} \label{eq:general_QP_formulation} \\
    -&\dot{\bm{x}}_R^{max} \le \dot{\bm{x}}_R \le \dot{\bm{x}}_R^{max} \nonumber 
    \\
    -&\Delta t_p \ \ddot{\bm{x}}_R^{max} \le \dot{\bm{x}}_R -\dot{\bm{x}}_{R_{0}} \le \Delta t_p \ \ddot{\bm{x}}_R^{max}
    \nonumber 
\end{align}
where $\bm{x}_r^{t_c}, \dot{\bm{x}}_r^{t_c} \in\mathbb{R}^{6}$ and $\bm{x}_o^{t_c}, \dot{\bm{x}}_o^{t_c}\in\mathbb{R}^{6}$ are the robot and object Cartesian poses and velocities at the catching instant.
The complete trajectories for the robot $\dot{\bm{x}}_R, {\bm{x}}_R \in\mathbb{R}^{6 T_p}$ 
are described until the estimated contact time $t_c$ (which is receding and obtained from the trajectory estimation detailed in Sec. \ref{sec:estimator}) and include the instantaneous poses of the robot $\bm{x}_r^i, \dot{\bm{x}}_r^i \in\mathbb{R}^{6}$ 
at the generic $i-$th time instant, in which $T_p = t_c/\Delta t_p$ is the total number of time samples until contact and $\Delta t_p$ is the sampling period of the planner.
We define:
\begin{equation} \label{eq:planner_vectors}
\dot{\bm{x}}_R = \begin{bmatrix}
        \dot{\bm{x}}^{i}_r\\
        \dot{\bm{x}}^{i+1}_r\\
        \hdots\\
        \dot{\bm{x}}^{i+T_p-1}_r\\
    \end{bmatrix} \hspace{-.1cm},
    {\bm{x}}_R = \begin{bmatrix}
        {\bm{x}}^{{i+1}}_r\\
        {\bm{x}}^{{i+2}}_r\\
        \hdots\\
        {\bm{x}}^{{i+T_p}}_r\\
    \end{bmatrix}\hspace{-.1cm},
     {\bm{x}}_{R_{0}} \hspace{-.1cm} = \hspace{-.1cm} \begin{bmatrix}
        {\bm{x}}^{{i}}_r \\
        {\bm{x}}^{{i+1}}_r \\
        \hdots \\
        {\bm{x}}^{i+T_p-1}_r
    \end{bmatrix}
    \hspace{-.15cm}
    \in\hspace{-.05cm}\mathbb{R}^{6 T_p}
\end{equation}
considering $ {\bm{x}}^{i+T_p}_r = \bm{x}_r^{t_c}$ and $\dot{\bm{x}}^{i+T_p-1}_r = \dot{\bm{x}}_r^{t_c}$.
Instead, ${\bm{x}}_{R_{0}}$ is defined analogously to $\bm{x}_R$ but shifted of one time sample to include the initial pose of the robot $\bm{x}^i_r$. $\bm{x}_R^{min},\bm{x}_R^{max}, \dot{\bm{x}}_R^{max}, \ddot{\bm{x}}_R^{max} \in\mathbb{R}^{6 T_p}$ are the boundaries for minimum and maximum position, velocity, and acceleration, respectively.
In particular, considering the velocities: 
\begin{equation} \label{eq:vel_lim_plan}
    \dot{\boldsymbol{x}}^{max}_R = \left[ \dot{\bm{x}}^{max}_r, \hdots, \dot{\bm{x}}^{max}_r \right]^T
\end{equation}
with
\begin{equation} \label{eq:plannerlimit}
    \dot{\bm{x}}^{max}_r =  \left[ \dot{x}^{max}_{lin}, \dot{x}^{max}_{lin}, \dot{x}^{max}_{lin}, \dot{x}^{max}_{ang}, \dot{x}^{max}_{ang}, \dot{x}^{max}_{ang} \right]^T
\end{equation} in which $ \dot{x}^{max}_{lin}, \dot{x}^{max}_{ang} \in \mathbb{R}$ are the limit values for linear and angular velocities, respectively.
The same structure holds for the position $\bm{x}_R^{min}, \bm{x}_R^{max}$ and accelerations limits $\ddot{\bm{x}}_R^{max}$.

By developing the equality constraint in \eqref{eq:general_QP_formulation} as a function of the variable $\dot{\bm{x}}_R$, we obtain
\begin{equation}
    \Delta t_p 
    \begin{bmatrix}
        \bm{I}_{6\times 6}, \hdots, \bm{I}_{6\times 6}
    \end{bmatrix}_{6 \times 6Tp} \dot{\bm{x}}_R = \bm{x}_o^{t_c} - \bm{x}_r^i,
\end{equation}
which constrains the trajectory to reach the estimated object's location at the final catching instant $t_c$, while the cost function accounts for VM by minimizing the relative final velocities.

In conclusion, QP problem \eqref{eq:general_QP_formulation} provides the optimal velocity trajectory that the robot should follow until the catching point $\dot{\bm{x}}_R^*$, which is numerically integrated to obtain $\bm{x}_R^*$ and sent to a middle layer responsible for fitting the trajectory at the same rate of the HQP controller (from the planner $\Delta t_p$ to the controller $\Delta t$ time period). This allows to reduce the sampling of the planned trajectories (hence the optimization's size) to improve computational speed and run in real time. 
Finally, the HQP controller receives the reference values (Fig. \ref{fig:framework}), the estimated object's data coming from the estimator is updated  (pose $\bm{x}_o^{t_c}$, velocity $\dot{\bm{x}}_o^{t_c}$ and time $t_c$) and the control sequence is generated. The process is repeated online, to exploit the increasing accuracy of the Kalman filter as more data about the object's trajectory is available.

\subsection{Post-catching strategy} \label{sec:poc}
The overall POC includes two parts: offline human stiffness learning and online trajectory planning and stiffness scaling. To generate the latter, the human arm physical variable stiffness (HVS) profiles in a single-axis free-falling object-catching task are encoded from offline data based on demonstrations (see the POC-offline learning block scheme of Fig. \ref{fig:framework} ).
The desired trajectory for POC, starting from the catching point, is planned online and is triggered when the impact force is higher than the threshold $F_{th}$. Meanwhile, the learned HVS is scaled to multiple-axis with respect to the relative distance from the catching point and sent to the HQP controller with the planned reference trajectory (see the POC-online block scheme of Fig. \ref{fig:framework} ). The details are explained in the following parts.

\subsubsection{Offline human stiffness learning:}\label{sec:stiff_learning} to clearly show the human stiffness learning process, we organize this section in the following two parts: human arm endpoint 
stiffness estimation and stiffness profile learning.

\subsubsection*{Human arm endpoint stiffness estimation:}
\begin{figure}[t]
    \centering
    \includegraphics[trim=0.3cm 0.3cm 0.3cm 0.1cm,clip,width=.85\columnwidth]{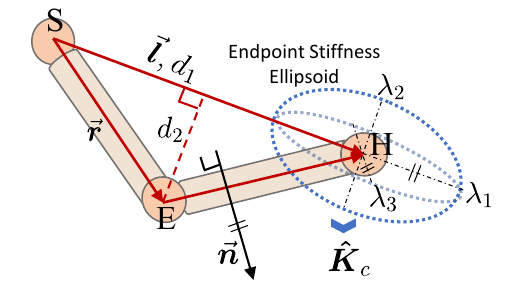} 
    \caption{Human arm configuration (3D) and parameters necessary to build the geometric model of the arm endpoint stiffness ellipsoid. S: shoulder; E: elbow; H: hand.
    }
    \label{fig:armstiff}
\end{figure}

the formulation in~\cite{wu2020intuitive} introduces the estimation of the human arm endpoint stiffness. As depicted in Fig.\ref{fig:armstiff}, the human arm is represented in 3D space, with the hand-forearm and upper arm segments forming a triangular configuration defined by the 3D position of the center of the shoulder, elbow, and hand at any non-singular arrangement. This triangular plane is similar to the arm plane, defined by the 3D position of the center of the shoulder, elbow, and wrist at any non-singular arrangement, to construct the swivel angle in~\cite{su2018online}.  
The vector $\vec{\boldsymbol{l}}\in\mathbb{R}^{3}$ extends from the shoulder joint's center to the hand's position, denoting the primary principal direction of the stiffness ellipsoid characterizing the human arm's endpoint.
Meanwhile, the vector $\vec{\boldsymbol{r}}\in\mathbb{R}^{3}$ connects the shoulder center to the elbow center. The minor principal axis direction, $\vec{\boldsymbol{n}}\in\mathbb{R}^{3}$, is established as orthogonal to the plane defined by the arm triangle. The calculation of the remaining principal axis of the stiffness ellipsoid, $\vec{\boldsymbol{m}}\in\mathbb{R}^{3}$, is determined based on the orthogonality constraints applied to the three principal axes.

Hence, the orthonormal matrix $\boldsymbol{V}\in\mathbb{R}^{3 \times 3}$ can be constructed as:
\begin{equation}
	\boldsymbol{V} = \Bigg [\frac{\vec{\boldsymbol{l}}}{\parallel \vec{\boldsymbol{l}} \parallel}, \frac{(\vec{\boldsymbol{r}} \times \vec{\boldsymbol{l}}) \times \vec{\boldsymbol{l}}}{\parallel (\vec{\boldsymbol{r}} \times \vec{\boldsymbol{l}}) \times \vec{\boldsymbol{l}} \parallel}, \frac{\vec{\boldsymbol{r}} \times \vec{\boldsymbol{l}}}{\parallel \vec{\boldsymbol{r}} \times \vec{\boldsymbol{l}} \parallel} \Bigg ].
	\label{eq:eigen_vectors}
\end{equation}

Based on the equation above, the stiffness becomes more elongated when the arm is extended, resulting in the hand moving further from the shoulder and it becomes more isotropic vice versa. Therefore, the proportion between the length of the median principal axis and the major principal axis of the stiffness ellipsoid exhibits an inverse relationship with the distance $d_1 \in\mathbb{R}^{+}$, extending from the hand position to the shoulder's center. Concurrently, the proportion of the minor principal axis's length to the major principal axis is postulated to be directly related to the distance $d_2 \in\mathbb{R}^{+}$, measured from the elbow center to the major principal axis. Furthermore, $d_1$ is inversely proportional to $d_2$. 
Based on the above observations, we assume $\frac{\lambda_{2}}{\lambda_{1}} = \frac{\alpha_1}{d_1}$ and $\frac{\lambda_{3}}{\lambda_{1}} = \alpha_2 \cdot d_2$, where $\lambda_{1} \in\mathbb{R}^{+}$, $\lambda_{2} \in\mathbb{R}^{+}$ and $\lambda_{3} \in\mathbb{R}^{+}$ represent the eigenvalues (i.e., the length) corresponding to the major, median and minor principal axes, respectively. $\alpha_1 \in\mathbb{R}$ and $\alpha_2 \in\mathbb{R}$ are scalar variables, and $d_1 = \parallel \vec{\boldsymbol{l}} \parallel$, $	d_2 = \vec{\boldsymbol{r}} \cdot \frac{\vec{\boldsymbol{m}}}{\| \vec{\boldsymbol{m}} \|}$.

In this model, synergistic muscle co-contractions are assumed to contribute to the endpoint stiffness ellipsoid's volume, expressed by the active component $A_{cc}\in\mathbb{R}$. $A_{cc}(p) = c_1 \cdot p + c_2$ is a linear relation with the muscle activation level $p\in\mathbb{R}$. 
The diagonal matrix $\boldsymbol{D}\in\mathbb{R}^{3 \times 3}$ is formed by the length of the principal axes, i.e., the eigenvalues,
\begin{equation}
	\boldsymbol{D} = A_{cc}(p) \cdot \boldsymbol{D}_s = A_{cc}(p) \cdot 
	\frac{\text{diag} (1, \quad \alpha_1 / d_1, \quad \alpha_2 d_2)}{ (1 \times \alpha_1 / d_1 \times \alpha_2 d_2)^{\frac{1}{3}}},
\label{eq:eigen_values}
\end{equation}
where $\lambda_{1}$ is set as $A_{cc}$, and $\boldsymbol{D}_s \in\mathbb{R}^{3 \times 3}$ represents the shape of the stiffness ellipsoid. Finally, the estimated endpoint stiffness matrix $\boldsymbol{\hat{K}}_c\in\mathbb{R}^{3 \times 3}$ is formulated by
\begin{equation}
	\boldsymbol{\hat{K}}_c = \boldsymbol{V}\boldsymbol{D}\boldsymbol{V}^T = \boldsymbol{V}A_{cc}(p)\boldsymbol{D}_s\boldsymbol{V}^T.
	\label{eq:stiffness_model}
\end{equation}
We use the average value of the four parameters identified by ~\cite{wu2020intuitive}: \resizebox{.9\columnwidth}{!}{$c_1=2033.325 $, $c_2=140.606 $, $\alpha_1=0.255$, $\alpha_2=2.815$}. 

\subsubsection*{Stiffness profile learning:} In the above subsection, the human arm endpoint stiffness matrix $\boldsymbol{\hat{K}}_c$ is estimated at each hand position, corresponding to each arm configuration, similar to~\cite{Wu2020aframe}. Here, we choose GMM to learn the relationship between the hand position and the corresponding stiffness profiles because it allows high dimensional vectors as input/output variables.     
Since GMM can not directly treat matrix (the translational stiffness matrix $\boldsymbol{\hat{K}}_c$) as output variable, the Cholesky decomposition is used to decompose it to a vector, which can be written as:
\begin{equation}
	\boldsymbol{\hat{K}}_c = \boldsymbol{L}^{T}\boldsymbol{L},
	\label{eq:stiffness_decom}
\end{equation}
where $\boldsymbol{L}\in\mathbb{R}^{3 \times 3}$ is a upper triangle matrix. 
The non-zero items of $\boldsymbol{L}$ are organized as a vector $\hat{\bm{L}}\in\mathbb{R}^{6}$, which is considered as the output variable ($\boldsymbol{\eta}^{\mathcal{O}}=\hat{\bm{L}}$) of the GMM. Since the arm stiffness depends on the hand position according to the used stiffness estimation model~\citep{wu2020intuitive}, the hand position should be the input variable.
However, considering the PRC process, instead of the hand position, we choose the relative distance from the catching point as the input variable ($\boldsymbol{\eta}^{\mathcal{I}}=\Delta d_h\in\mathbb{R}^{+}$).
A GMM is defined as a linear superposition of several Gaussian distributions:
\begin{equation}
	p(\boldsymbol{\xi}) = \sum_{k=1}^{K}\pi_k\mathcal{N}(\boldsymbol{\xi}|\boldsymbol{\mu}_k,\boldsymbol{\Sigma}_k),
\end{equation}
where $\boldsymbol{\xi}=[ \Delta d_h, \, \hat{\bm{L}}^T]^T \in\mathbb{R}^{d}$ and $p(\boldsymbol{\xi}) \in\mathbb{R}$ represent the vector of variables (input and output variables) and joint probability distribution, respectively. $\pi_k \in\mathbb{R}$, $\boldsymbol{\mu}_k\in\mathbb{R}^{d}$ and $\boldsymbol{\Sigma}_k\in\mathbb{R}^{d \times d}$ are the prior probability, mean and covariance of the $k$-th Gaussian component, respectively. $K\in\mathbb{Z}^+$ is the number of Gaussian distributions.
The parameters of GMM can be estimated by Expectation-Maximization (EM) algorithm~\citep{bishop2006pattern} with an offline training process that uses the demonstrations. 

To make the representation easier to understand, we give the following definition:  $\boldsymbol{\eta}^{\mathcal{I}}=\Delta d_h$ and $\boldsymbol{\eta}^{\mathcal{O}}=\hat{\bm{L}}$ which respectively denote the input and output variables on which the training is performed.

Based on these, the generic data point $\boldsymbol{\eta} \in \mathbb{R}^{d}$, the mean $\boldsymbol{\mu}_k$ and covariance $\boldsymbol{\Sigma}_k$ of the the $k$-th Gaussian component can be written as: 
\begin{equation}
	\boldsymbol{\eta} =
	\begin{bmatrix}
		\boldsymbol{\eta}^{\mathcal{I}} \\
		\boldsymbol{\eta}^{\mathcal{O}}
	\end{bmatrix}=
        \begin{bmatrix}
		\Delta d_h \\
		\hat{\bm{L}}
	\end{bmatrix},
	\boldsymbol{\mu}_k = 
	\begin{bmatrix}
		\boldsymbol{\mu}_k^{\mathcal{I}} \\
		\boldsymbol{\mu}_k^{\mathcal{O}}
	\end{bmatrix},
	\boldsymbol{\Sigma}_k = 
	\begin{bmatrix}
		\boldsymbol{\Sigma}_k^{\mathcal{I}\mathcal{I}} & \boldsymbol{\Sigma}_k^{\mathcal{I}\mathcal{O}} \\
		\boldsymbol{\Sigma}_k^{\mathcal{O}\mathcal{I}} & \boldsymbol{\Sigma}_k^{\mathcal{O}\mathcal{O}} 
	\end{bmatrix}. 
\end{equation}
Given an input variable $\boldsymbol{\eta}^{\mathcal{I}}$, the best estimation of output $\boldsymbol{\hat{\eta}}^{\mathcal{O}}$ is the mean $\boldsymbol{\hat{\mu}}$ of the conditional
probability distribution $\boldsymbol{\hat{\eta}}^\mathcal{O}|\boldsymbol{\eta}^\mathcal{I} \sim \mathcal{N}(\boldsymbol{\hat{\mu}}, \boldsymbol{\hat{\Sigma}})$, which can be obtained by \citep{huang2019kernelized}:
\begin{equation}
\label{eq:GMR}
	\boldsymbol{\hat{\mu}} = \mathbb{E}(\boldsymbol{\hat{\eta}}^\mathcal{O}|\boldsymbol{\eta}^\mathcal{I}) = \sum_{k=1}^{K}h_k(\boldsymbol{\eta}^\mathcal{I})\boldsymbol{\mu}_k(\boldsymbol{\eta}^\mathcal{I}),
\end{equation}
where
\begin{equation}
	h_k(\boldsymbol{\eta}^\mathcal{I}) = \frac{\pi_k\mathcal{N}(\boldsymbol{\eta}^\mathcal{I}|\boldsymbol{\mu}_k^{\mathcal{I}},\boldsymbol{\Sigma}_k^{\mathcal{I}\mathcal{I}})}{\sum_{j=1}^K\pi_j\mathcal{N}(\boldsymbol{\eta}^\mathcal{I}|\boldsymbol{\mu}_j^{\mathcal{I}},\boldsymbol{\Sigma}_j^{\mathcal{I}\mathcal{I}})},
\end{equation}
\begin{equation}
	\boldsymbol{\mu}_k(\boldsymbol{\eta}^\mathcal{I}) = \boldsymbol{\mu}_k^\mathcal{O}+\boldsymbol{\Sigma}_k^{\mathcal{O}\mathcal{I}}(\boldsymbol{\Sigma}_k^{\mathcal{I}\mathcal{I}})^{-1}(\boldsymbol{\eta}^\mathcal{I}-\boldsymbol{\mu}_k^{\mathcal{I}}).\\
\end{equation}

Here, the GMM model is trained offline based on collected human demonstrations, whose results are presented in Sec.\ref{subsec:human_demon}. The relative distance from the catching point $\Delta d_h$ and the stiffness vector $\hat{\bm{L}}$ are chosen as the input and output variables, respectively. Therefore, the stiffness depends on the location with respect to the catching point. The state-dependent design can adapt to randomly thrown objects since it is triggered by the impact force and is updated by the relative distance.

\begin{figure}[t]
    \centering
    \includegraphics[trim=0.3cm 0.3cm 0.3cm 0.3cm,clip,width=.85\columnwidth]{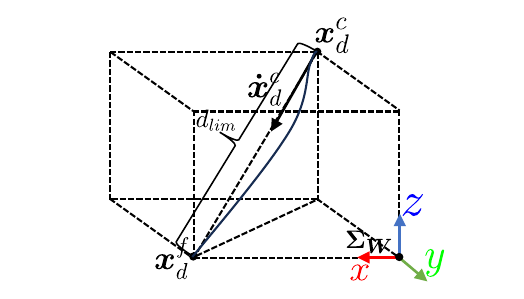} 
    \caption{Planned trajectory (solid curve) for POC in 3D space.
    }
    \label{fig:poc_tra}
\end{figure}

\subsubsection{Online trajectory planning and stiffness scaling:}\label{sec:poc_online}

to transfer the learned human behavior in catching free-falling objects to the randomly thrown objects in POC, the trajectory should be planned online in a short time, still attaining the task space constraints dictated by the experimental setup.
Meanwhile, the learned single-axis HVS should be mapped and scaled online to the multi-axis scenario. 

Considering we just have the requirements at position and velocity level for the initial and final points
of the POC trajectory, we choose a third-order polynomial to plan the trajectory for POC, which ensures the continuity of position, velocity, and acceleration profiles. As mentioned, only the position part of the EE is planned, while the orientation does not change. The initial conditions are the last reference position and velocity obtained from the VM planner in \eqref{eq:general_QP_formulation}, thus for legibility, we define: 
$\bm{x}_d^c = {\bm{x}}^{i+T_p}_r$ for the desired position and $\dot{\bm{x}}_d^c = \dot{\bm{x}}^{i+T_p-1}_r$ for the velocity.
$d_{lim}\in \mathbb{R}$ is the maximum moving distance for the POC, dictated by the feasible task space of the robot, and
$\bm{x}_d^f\in \mathbb{R}^6$ is the final point of the POC trajectory, which can be calculated by projecting $d_{lim}$ along the direction of $\dot{\bm{x}}_d^c$:
\begin{equation}\label{eq:final_point}
\left\{\begin{matrix}
x_d^f=x_d^c+(d_{lim}^{xy}\dot{x}_d^c/V_{d,xy}^c),\\ 
y_d^f=y_d^c+(d_{lim}^{xy}\dot{y}_d^c/V_{d,xy}^c),\\ 
z_d^f= z_d^c+d_{lim}^{z},\\
d_{lim}^{z}=d_{lim}\dot{z}_d^c/V_d^c,
\end{matrix}\right. 
\end{equation} 
where $x_d^c,y_d^c,z_d^c$ and $x_d^f,y_d^f,z_d^f$ are the position terms of $\bm{x}_d^c$ and $\bm{x}_d^f$, respectively. $\dot{x}_d^c,\dot{y}_d^c,\dot{z}_d^c$ are the translational velocities of $\dot{\bm{x}}_d^c$. $d_{lim}^{z},d_{lim}^{xy}=\sqrt{({d}_{lim})^{2}-({d_{lim}^{z}})^{2}}$ are the projections of $d_{lim}$ in $z-$axes and $xy-$plane, respectively. $V_d^c=\sqrt{(\dot{x}_d^c)^{2}+(\dot{y}_d^c)^{2}+(\dot{z}_d^c)^{2}}$ is the absolute value of the translational velocity of $\dot{\bm{x}}_d^c$, and $V_{d,xy}^c=\sqrt{(\dot{x}_d^c)^{2}+(\dot{y}_d^c)^{2}}$ is the absolute value of the translational velocity of $\dot{\bm{x}}_d^c$ in the $xy-$plane. The detailed relationship between these variables for POC trajectory planning can be found in Fig. \ref{fig:poc_tra}.
Since $\dot{\bm{x}}_d^f=\bm{0}$, the period $\Delta t_{poc}$ to obtain a kinematically feasible trajectory for POC can be chosen as $\Delta t_{poc}~\geqslant~{max(\dot{\bm{x}}_d^c/\ddot{\bm{x}}^{max})}$.

Once the impact force triggers the POC, the overall trajectory is generated by the above planner. The relative distance from the catching point can be scaled as:
\begin{equation}
	\Delta d_h=\frac{\left \| \boldsymbol{x}_a-\boldsymbol{x}_d^c\right \|}{d_{lim}}d_h,
	\label{eq:distance_scale}
\end{equation}
where $\bm{x}_a \in \mathbb{R}^6$ is the actual EE's pose at the current time, whereas $d_h$ is the human movement length from the catching point to the endpoint. The current relative distance from the catching point $\Delta d_h$ is sent to \eqref{eq:GMR} as input variable $\boldsymbol{\eta}^{\mathcal{I}},$ to calculate the corresponding output variable $\hat{\bm{L}}$. After that, the learned human physical stiffness $\bm{K}_d$ (namely, $\boldsymbol{\hat{K}}_c$ ) can be obtained by \eqref{eq:stiffness_decom}. Finally, the control gain $\bm{K}_p$ for the HQP controller (as visible in Fig. \ref{fig:framework}) can be scaled as:
 \begin{equation}
	\bm{K}_p=\frac{\bm{K}_d}{{K}_d^{max}}{K_p^{max}},
	\label{eq:stiffness_scale}
\end{equation}
where the ${K}_d^{max},K_p^{max} \in\mathbb{R}$ are the maximum values allowed for $\bm{K}_d$ and $\bm{K}_p$ respectively, and depend on the specific application. 
The overall process of online trajectory and stiffness generation and scaling is described in Algorithm \ref{alg:stiff}. 

\begin{algorithm}[t]
\caption{Online stiffness generation and scaling} 
\label{alg:stiff}
\hspace*{0.02in} {\bf Input:} 
$\left \| \bm{F} \right \|, \bm{x}_{d}^c, \dot{\bm{x}}_d^c, d_{lim}, \Delta t_{poc}, F_{th}$\\
\hspace*{0.02in} {\bf Output:} 
$\bm{x}_d, \dot{\bm{x}}_d, \bm{K}_p$
\begin{algorithmic}[1]
\If{ $\left \| \bm{F} \right \|> F_{th}$ }
        \State calculate $\bm{x}_d^f$ by \eqref{eq:final_point} ,
        \State plan the overall POC trajectory,
        \While{$t \leqslant \Delta t_{poc}$}
               \State obtain current desired trajectory $\bm{x}_d, \dot{\bm{x}}_d$,
               \State calculate relative distance by \eqref{eq:distance_scale},
               \State get stiffness vector $\hat{\bm{L}}$ by \eqref{eq:GMR},
               \State obtain learned HVS by \eqref{eq:stiffness_decom},
               \State calculate control gain $\bm{K}_p$ by \eqref{eq:stiffness_scale}.
        \EndWhile
        \State $t\leftarrow t+\Delta t $.
\EndIf
\end{algorithmic}
\end{algorithm}

\subsection{Hierarchical control} \label{sec:method_hqp}

\subsubsection{Motivations:}
we exploit a well-studied technique in the field of hierarchical control, namely HQP, mainly to i) enable robot multitasking and secondary RMM, ii) ensure continuity at the impact by enforcing robot kinematic/dynamic limits.
The formulation is similar to existing works, the only addition consists in formulating a secondary layer to perform RMM via the maximization of the DIM ellipsoid along the impact axis, dictated by the one-dimensional impact model.

The first motivation i) is related to the large and quick impulsive force generated at the EE, which might lead to severe damages to the robot and its equipment. 
Hence, the goal is to optimally redistribute joints' activation by adjusting the robot's redundant configuration. 

The other motivation ii),
is the generation of accurate yet feasible motion. Indeed, in dealing with fast-flying objects, a prompt and accurate controller is necessary to successfully track a rapidly-varying reference in a short time window.
However, feasibility of the generated control output must be ensured, which is possible thanks to the constrained optimization that accounts for physical and practical limitations. This allows to maintain the same controller in the transition from PRC to POC, exploiting its stability features ensured by the convexity of the QP problem and its robustness to external disturbances~\citep{Escande}, as it will be discussed in Sec.~\ref{sec:stability}.

\subsubsection{Hierarchical structure:}

Considering the generic HQP-based control problem defined in \cite{augmentedTassi}, a Stack of Tasks (SoT) is formulated to impose a \textit{strict hierarchy}, in which $N_p\in\mathbb{N}$ is the total number of hierarchical layers in the SoT, and  $k \in \{1, \dots N_p\}$ are the levels of priority with decreasing importance from $1$ to $N p$.
This ensures that the solution found at level $k$ is strictly enforced at the lower priority $k+1$, thanks to the optimality condition defined at each QP solution in \cite{Kanoun}.

\subsubsection{Closed-loop inverse kinematics:}

The first task necessary in the SoT is the Closed-loop Inverse Kinematics (CLIK), which closes the loop on the Cartesian EE error, to achieve trajectory tracking both with respect to the desired pose and velocity trajectories $\bm{x}_d, \dot{\bm{x}}_d \in \mathbb{R}^6$
\begin{equation} \label{eq:min_kyn_clik}
    \min_{\dot{\bm{q}}} \,\| \bm{J}\dot{\bm{q}}-( \bm{K_v} ( \dot{\bm{x}}_d -\dot{\bm{x}}_a ) + \bm{K_p}\,(\bm{x}_d-\bm{x}_a) ) \|^2
\end{equation}
where $\bm{x}_a, \,\bm{x}_d \in \mathbb{R}^6$ are the actual and desired EE trajectories, respectively; $\dot{\bm{x}}_a, \,\dot{\bm{x}}_d \in \mathbb{R}^6$ are respectively the actual and desired velocities, while $\bm{K_p}, \bm{K_v} \in \mathbb{R}^{m\times m}$ are the positive-definite diagonal gain matrices responsible for the accuracy of position and velocity trajectories' tracking.
$\dot{\bm{q}} \in \mathbb{R}^n$ are the joint velocities and $\bm{J} \in \mathbb{R}^{m\times n}$ is the Jacobian matrix.

\subsubsection{Minimum Reflected Mass:} \label{sec:mrm}

Overall, the hierarchical controller can simultaneously handle: 
i) motion tracking accuracy, for successfully catching the flying object, ii) both joint and task space constraints, iii) the relationship between the impulsive force at the EE and its configuration.
Indeed, this relationship has been addressed in Sec. \ref{sec:impact-model}, since to achieve RMM, the goal is to minimize the impact force \eqref{eq:impact_relationship} thus equivalently minimize \eqref{eq:rmm_def}.
This is proportional to the \textit{dynamic impact ellipsoid} formulated in \cite{Walker1994} as:
\begin{equation} \label{eq:dyn_impact_ellipsoid}
    \bm{u}^T \bm{J}(\bm{q}) \bm{M}(\bm{q})^{-2} \bm{J}(\bm{q})^T \bm{u} \leq 1,
\end{equation}
where $\bm{u} \in \mathbb{R}^6$ is the direction of the contact forces. This ellipsoid represents how the robot configuration is susceptible to impact forces in space along a specific direction.
Our goal is to maximize the robot's capability of withstanding an impact at the EE, along the impact's direction. However, in doing so, the primary task of accurately and rapidly reaching the contact location should not be affected, therefore having to rely on a strict hierarchical structure.

To this purpose, a quantitative index is obtained from the \textit{Dynamic Impact Measure} (DIM) \citep{Walker1994}, which is proportional to the volume of the \textit{dynamic impact ellipsoid} \eqref{eq:dyn_impact_ellipsoid}. We rewrite this index by making explicit the impulse direction $\bm{u}$ as
\begin{equation} \label{eq:dm_index}
w_{f_{d}}(\bm{q}) :=
\bm{u}^T \bm{J}(\bm{q})^{+T} \bm{M}(\bm{q}) \,\bm{M}(\bm{q})^T\,\bm{J}(\bm{q})^+ \bm{u}.
\end{equation}
with an appropriate choice of the pseudoinverse $\bm{J}(\bm{q})^+$ \citep{Walker1994}.
Maximizing this index along the direction of the impact results in identifying the redundancy configuration that allows to better withstand an impact at the EE.
Indeed, as explained in \cite{Walker1994}, the \textit{dynamic impact ellipsoid} indicates the directions in which the robot is dynamically stiff, as opposed to maneuverable, recalling the dual relationship between the manipulability and manipulating force ellipsoids in the kinematics-only analysis \citep{Walker2}.
It is interesting to study the catching behaviour when optimizing for stiffness as opposed to manipulability, hence the activation of DIM is further analyzed in the experiments.

The problem of maximizing $w_{f_{d}}(\bm{q})$ is defined as
\begin{equation} \label{eq:max_wfd}
    \min_{\dot{\bm{{q}}}} \ - w_{f_{d}}(\bm{q}),
\end{equation}
which is a nonlinear optimization problem in $\bm{q}$, whose linearization and formulation in QP form is obtained as in \cite{dufour2020maximizing, Tassi_Gholami2021}, assuming the second-order Taylor expansion is accurate enough to describe the nonlinearities
\begin{equation}
    w_{f_{d}}(\bm{q}) =  w_{f_{d}}(\bm{q}_{t-\Delta t}) + \Delta t\, (\bm{\bm{\nabla}} w_{f_{d}})^T \bm{\dot{q}} + \frac{1}{2}\,  \Delta t^2\, \bm{\dot{q}}^T \,\bm{H}_w\,  \bm{\dot{q}},
\end{equation}
where $\Delta t$ is the control period and $\bm{\nabla} w_{f_{d}}\in\mathbb{R}^n$ is the gradient vector of $w_{f_{d}}$ and $\bm{H}_w\in\mathbb{R}^{n \times n}$ is the Hessian matrix.
An equivalent optimization problem is proposed in \cite{dufour2020maximizing} to avoid the use of the Hessian matrix which is often computationally intesive:
\begin{equation}
    \label{eq:min_mi_index}
    \min_{\dot{\bm{q}}}  \frac{1}{2}\,  \Delta t^2\, \dot{\bm{q}}^T\, \bm{\nabla} w_{f_{d}}\, (\bm{\nabla} w_{f_{d}})^T \dot{\bm{q}} - w_{f_{d}}\, \Delta t\, (\bm{\nabla} w_{f_{d}})^T \dot{\bm{q}}.
\end{equation}

In conclusion, \eqref{eq:min_mi_index} is used as a secondary layer in the SoT, to instantaneously minimize the reflected mass at the EE, as if the impact is to occur at the following time instant. This ensures greater robustness against contact timing.
In this sense, a similar approach is used in \cite{YuquanWang_new}, where an impact-aware QP problem is formulated to instantaneously account for potential impacts, by constraining the feasible set based on the hardware properties. However, this does not provide the advantages of strict hierarchies and in the POC phase the impact-aware constraints only act immediately after the impact, whereas no actual longer-term policy is considered to dissipate the catching-induced energy (being the grasping prehensile).

\subsubsection{Constraints:}

The constraints that regulate the HQP problem are defined at position, velocity, and acceleration levels based on the physical limits of the robot's actuators:
\begin{gather}
        \bm{q}_{min} \leq \bm{q}_{t-\Delta t} +\dot{\bm{q}} \ \Delta t \leq \bm{q}_{max} \nonumber \\
        \dot{\bm{q}}_{min} \leq \dot{\bm{q}} \leq \dot{\bm{q}}_{max} \label{eq:qp_constraints}\\
        \ddot{\bm{q}}_{min} \leq \frac{\dot{\bm{q}} -\dot{\bm{q}}_{t-\Delta t}}{\Delta t} \leq \ddot{\bm{q}}_{max} \nonumber
\end{gather}
where $ \bm{q}_{t-\Delta t} \in \mathbb{R}^n $ is the optimal value identified at the previous time instant
whereas $\bm{q}_{min}, \dot{\bm{q}}_{min}, \ddot{\bm{q}}_{min} \in \mathbb{R}^n$ and $\bm{q}_{max}, \dot{\bm{q}}_{max}, \ddot{\bm{q}}_{max} \in \mathbb{R}^n$ are the joint limits in position, velocity, and acceleration, respectively.

\subsubsection{Stability:} \label{sec:stability}

Being written in QP form, the HQP problem formulated above is convex and thus proven stable via the addition of the regularization term $\min_{\dot{\bm{{q}}}} \| \dot{\bm{{q}}} \|^2 $ in the SoT \citep{Kanoun, Escande}, which is added in each HQP problem studied, as a last priority in the SoT.
Assuming the initial robot configuration to be reasonably far from singularities, the regularization term also acts as singularity avoidance, penalizing singular configurations.
Additionally, provided the constraints represent a feasible set, stability is ensured also in the case in which the initial conditions lie outside of the feasible region, resulting in a continuous transition towards such region, as demonstrated in \cite{Escande}.

Given the high dynamism of the task at hand and the short time window in which the catch has to occur, sharp velocity transitions are necessarily generated by the PRC planner upon object's detection. However, these still respect the limits imposed via the constraints in \eqref{eq:general_QP_formulation}, and are further filtered by the controller, whose constraints \eqref{eq:qp_constraints} allow the generation of continuous and stable joint trajectories.
After the transition between PRC and POC instead, the Cartesian trajectories and stiffness references are now generated by the POC module in Fig.~\ref{fig:framework}, but the controller remains the same, hence preserving stability.

\subsection{Joint Impedance Control} \label{sec:jic}

Being our main focus to i) analyze the impact forces and ii) replicate a specific impedance behavior in the POC phase, an impedance controller is necessary. Hence, a low-level decentralized joint impedance controller is used to generate the desired torque profiles from the optimal joint velocities obtained from the HQP (Fig. \ref{fig:framework}):
\begin{equation} \label{eq:joint_imp_ctrl}
    \bm{\tau} = \bm{K_{q_{d}}} (\dot{\bm{q}}^* - \dot{\bm{q}}_a) + \bm{K_{q_{p}}} (\bm{q^* - q}_a)+ \boldsymbol{C}(\bm{q},\dot{\bm{q}})  + \bm{g}(\bm{q}),
\end{equation}
where, $\bm{q}^*, \dot{\bm{q}}^* \in \mathbb{R}^n$ are the desired optimal joints' position and velocity obtained from the HQP, $\bm{q}_a, \dot{\bm{q}}_a \in \mathbb{R}^n$ are the actual joint positions and velocities respectively, while $\bm{K_{q_{p}}} \in \mathbb{R}^{n\times n}$ and $\bm{K_{q_{d}}} \in \mathbb{R}^{n\times n}$ are the positive definite joint stiffness and damping matrices, respectively.

It is important to note, that the Cartesian stiffness behaviour of the robot in POC would slightly diverge from the learnt human stiffness values, in favor of satisfying additional tasks (in the SoT) and constraints. However, we believe that the actual profile of the stiffness in different phases of the POC plays a more important role than its absolute value in such dynamic tasks \citep{ajoudani2012tele}, since it regulates the trade-off between tracking and energy dissipation. 
\section{Experimental Setup}\label{sec:experimental-setup}

In this section, we describe the overall setup and provide details about the human demonstrations, the perception and trajectory estimation, and the parameters used in the robot's control. This section also defines the metrics used to compare the methods. We first conduct a thorough investigation in a single dimension (Sec. \ref{sec:exp1D}), where the object is free-falling vertically, only affected by gravity. The motivations for this choice are twofold. 

Firstly, this is essential to thoroughly test and accurately validate the effects that the addition/removal of each of the component that characterizes the proposed framework has on the multiple metrics under study. 
Indeed, to perform a proper comparison, the goal is to first remove any unpredictability in the throwing of the object so that the impact forces are repeatable. Therefore, instead of considering the scenario in which the object is thrown by a human, which inherently leads to a series of random components in terms of trajectories, impact time, impact position and forces, it is preferable to simply drop the object from a fixed height. In this way, repeatability is ensured, being the free-falling object only affected by the gravitational pull.

Secondly, this serves as a comparison with our previous work, in which only the single-dimensional case was analyzed. The aim is to highlight the substantial differences of this work and to comprehend its improvements.

Eventually, we generalize the proposed framework in space by extending the problem's dimension and validating the proposed catching architecture in the case where the human throws the object (Sec. \ref{sec:exp3D}).

Since our goal is to analyze a generic catching scenario, we try to replicate the impact of an object onto a generic flat surface to ensure repeatable and consistent results. To do so, we impose the orientation of the EE parallel to the ground at the catching point. Being the EE designed as a flat basket (Fig.~\ref{fig:1d-exp-setup}) together with a ball-shaped flying object, ensures that the relative orientation at the contact instant remains fixed, providing a consistent baseline. Indeed, the forces generated by the impact between two generic objects strongly depend on their relative orientations at the impact, hence, with the proposed design, we remove this source of variability that would amplify when increasing the dimensionality of the task and across multiple trials.
Besides, we avoid additional dynamic effects of self-induced EE inertial forces.

Despite this choice, both the proposed optimal planner and controller currently account for generic time-varying orientations. Indeed, by changing the catching reference in the proposed optimal planning problem \eqref{eq:general_QP_formulation}, it is possible to generate non-flat final EE configurations, without restricting the framework's theoretical applicability. 
This meaning that the instantaneous/final orientations of the EE could be chosen as horizontal (as we did due to the shape of the EE), or based on other aspects, such as considering the estimated orientation of the flying object at the catching point, obtained via perception.
In this way, the flat area of the EE would always hit the object perpendicularly with respect to the object's direction, theoretically improving the repeatability of the measured impact forces at the EE throughout multiple experiments. Eventually, the POC planner would drive the EE's orientation from the catching instant to a final horizontal one, to avoid dropping the object.
We avoid this, as mentioned above, only to maintain a straightforward comparison between the single- and multi-axis scenarios and to avoid adding noise to the already short impact peak, resulting from aspects such as self-induced EE oscillations due to its inertia.

However, the generic nature of the proposed formulation with respect to the choice of EE orientation, is validated in the dimensionality extension experiments of Sec.~\ref{sec:exp3D}, where the the direction of the object is not aligned with the EE's $z$-axis (Fig.~\ref{fig:1d-exp-setup}). 
In this case, based on the aforementioned assumptions,
the choice of horizontal EE only affects the amount of
RMM achieved at the EE (since the DIM optimization uses the estimated object's direction at the catching point in \eqref{eq:dm_index}, while the EE's orientation is instead horizontal), which is sub-optimal, however, it does not affect the PRC planner hence the success of the catch. Therefore, this can be considered as a worst-case scenario, that can only improve when the estimated direction of the object at the catching instant is provided as a reference in \eqref{eq:general_QP_formulation} and therefore the catching EE's orientation is normal with respect to the object's direction.

While non-flat EE configurations are not useful in this specific study, due to our main focus on impact-analysis rather than catching method, 
they become of interest in e.g., prehensile catching scenarios, where the relative catching orientation is fundamental. Indeed, this strategy could contribute to the reduction of impact forces and prevent the common issue of bouncing of the object out of the EE before grasp's closure, in applications similar to \cite{kim2014catching, softcat2016tro, Bauml2010hirzinger, sato2020high}.
Clearly, considering the particularly complex nature of the catching task, no speculation can be made a priori and further studies are necessary to properly validate the theoretical feasibility of the framework, showing its applicability to other domains. Indeed, further issues would arise in this case, such as: accurately tracking spinning objects and estimating their trajectories, induced torsional effects which would affect both catching success and impact-minimization, grasp timing and catching success.

\subsection{Human demonstrations data collection}\label{subsec:human_demon}

An experimental investigation on a healthy human subject allowed to analyze the human-compliant behavior in POC.
\subsubsection*{Setup:}
the subject was required to wear a Lycra suit covered with markers, tracked online by the OptiTrack to collect the human upper-body motion data.
EMG sensors were placed on the subject's biceps to record muscle activity as they are the most activated muscles in the catching task \citep{humanCatching2021}.
The data acquisition and synchronization of all the sensors are managed using Robot Operating System (ROS) environment at 50 Hz. 

\subsubsection*{Protocol:}
the subject was required to perform $9$ catching tasks using both arms, according to the following protocol: at the beginning of each trial, the subject was asked to maintain his feet steady, with the arms placed forward at ninety degrees to the hips, waiting for the box. An assistant in front of the subject released a $5~kg$ box at a fixed height of $1~m$ (around eye level) from the subject's arms (see the offline learning part in Fig. 2). 
In addition, according to previous neuroscience research on human behavior in intercepting a free-falling target in~\citep{lacquaniti1989role,zago2009visuo}, the human's upper limbs EMG activity and corresponding impedance actively changed to adapt to different momentum (mass times velocity) of the object at the contact instant.
To obtain a proper contact momentum for the subject, we used a similar setup with respect to a recent neuroscience study~\citep{humanCatching2021}, which investigated the human response to a free-falling box weighing 2.5\% and 5\% of the subject's mass, released at the height of the subject's eyes. Similarly, in our case, the weight of the box was set around 5.2\% of the subject's weight (95.85kg, i.e., squat payload), and it was unknown to the subject.
To link the human's payload to the object's weight and obtain a one-to-one relation in the transfer to the robot, we consider the human payload-to-weight ratio of \cite{falch2023association}, whose average among resistance-trained individuals is reported as 1.68 for males and 1.35 for females, in a squat-like motion. We limit this ratio to 1 for safety purposes, considering the dynamic nature of the catching task.

\subsubsection*{Results:}
\begin{figure}
    \centering
    \includegraphics[width=.95\linewidth]{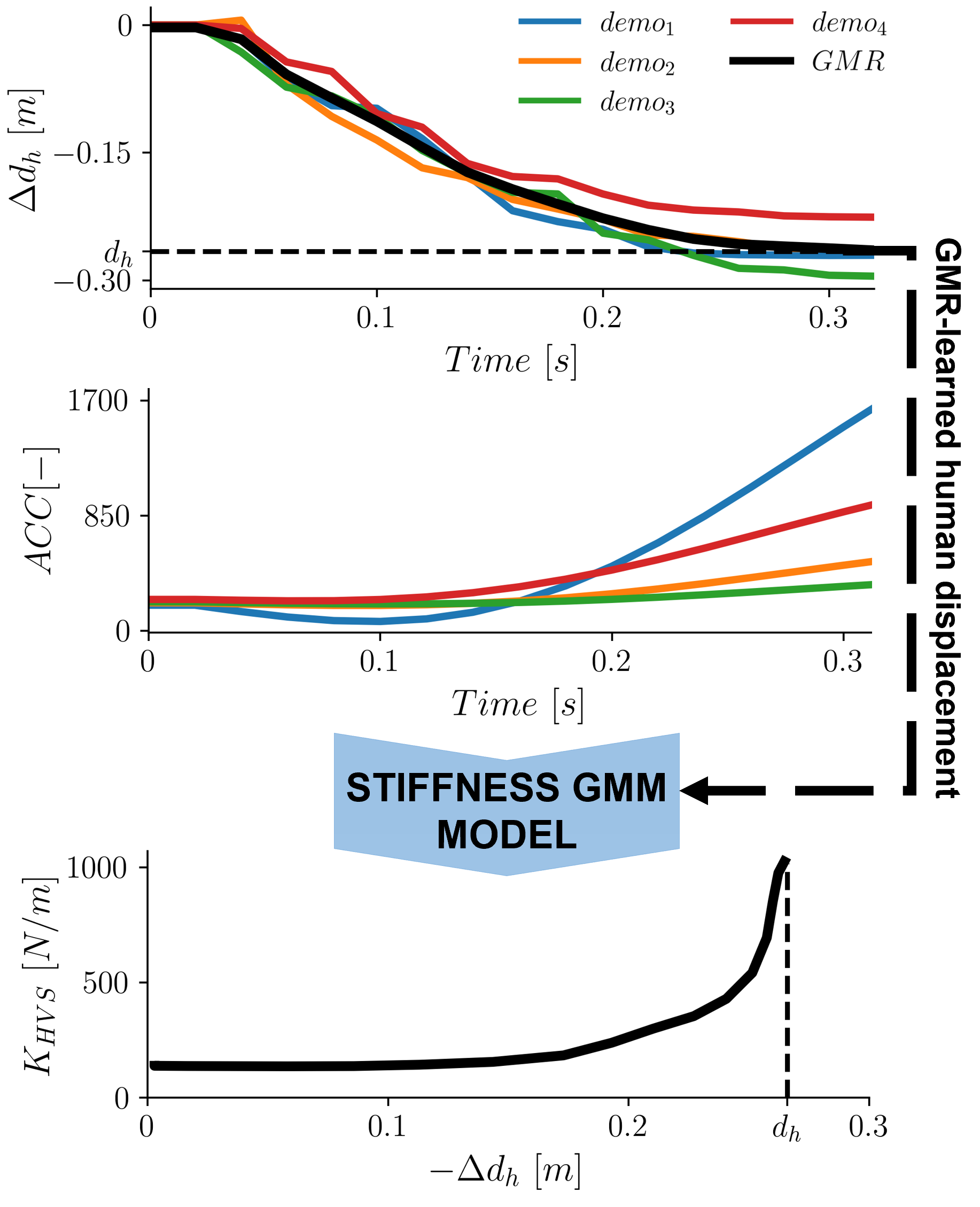}
    \caption{Human arm stiffness profile ($z-$axis) learned from demonstrations. It ranges from the moment the human touches the object until the lowest point in the trajectory. The human hand displacement $\Delta d_h$ obtained during catching and the $ACC$ generate a GMM model for the variable stiffness. At the same time, the displacements are learned by a GMR algorithm and then fed to the GMM model to obtain the $K_{HVS}$.
    }
    \label{fig:hvs}
    \vspace{-1.5mm}
\end{figure}

four demonstrations were selected based on the consistency of human movement, and the mean value of both arms was used to construct the stiffness (see Sec. \ref{sec:stiff_learning}). Specifically, the mean value of EMG sensors represents the muscle activation level $p$, which is used to calculate the $A_{cc}$ (see Fig.\ref{fig:hvs}). Here, we set the number $K=8$ of Gaussian distributions in GMM based on experience, which can be optimized by integrating the Bayesian Information Criterion~\citep{calinon2007learning} in future research.
Besides, the position of the shoulder, elbow, and hand are used to formulate the configuration-related parameters $\bm{V},\bm{D}_s$. Then, the stiffness is constructed by \eqref{eq:stiffness_model}. 
In the offline GMM learning process, the relative distance from the catching point $\Delta d_h$ and the stiffness vector $\hat{\bm{L}}$ are set as input $\boldsymbol{\eta}^{\mathcal{I}}$ and output $\boldsymbol{\eta}^{\mathcal{O}}$ variables. Although the whole translational stiffness matrix is recorded and learned, we consider the stiffness in the $z-$axis most relevant in the free-falling catching task.       

To obtain the general human movement length $d_h$, the GMM/GMR is used to learn the trajectory profile. The result is presented in Fig.~\ref{fig:hvs}, where the learned human demonstration had a displacement of $d_h= -0.27~m$ from the initial position in $0.32~s$.
Finally, we used the GMR result of $\Delta d_h$ as input to the previously trained stiffness GMM model and reconstructed the stiffness by \eqref{eq:stiffness_decom}. The learned HVS in the $z-$direction with respect to the $-\Delta d_h$ is presented in Fig. \ref{fig:hvs}. Specifically, the value of HVS increases gradually from the catching point (lower than $150~N/m$) to the endpoint (higher than $1000~N/m$). This learned behavior is consistent with our expectations and previous research ~\citep{ajoudani2012tele,yantro2024}: a soft behavior at the start to absorb the impact, followed by a stiffness increase to manipulate the object accurately. Furthermore, given the real-time relative distance from the catching point in the task space, we apply the learned human stiffness profile to the robotic arm during POC as described in Sec.\ref{sec:stiff_learning} to dissipate contact energy, avoid bouncing, and ensure a successful catch of the object.

\subsection{Perception and trajectory estimation}
Accurate trajectory tracking and prediction of flying objects can be improved using high frame rate cameras that have fast shutters and reduce motion blur - but which then require brightly lit environments. The fixed shutter photon integration period and the computational cost demanded by processing each frame introduces significant latency (tens to hundreds of ms) in a vision pipeline.
The event-camera~\citep{gallego2020event} is a recently developed alternative, without a shutter, which produces asynchronous and sparse data. Each pixel contains independent circuitry and when the level of illumination changes beyond a threshold, an \textit{event} is generated for the single pixel location. Without frame rate, the problem of motion blur becomes insignificant, while small packets of events (much less data than a full image frame) can be transferred and processed with millisecond latency. A high dynamic range allows event-camera use in both well- and dimly-lit conditions. Despite the strong potential, event-based perception is still a developing field, and robust algorithms for specific robotic tasks are not yet as widely available or as easily accessible and integrable in robotic applications as traditional computer vision. 

In the loop, event-based vision for robot control includes quadrotor dodging~\citep{falanga2020dynamic} with 3.5 ms latency when fused with an inertial measurement unit and 60 ms latency using a vision-only deep learning approach~\citep{sanket2020dodging}. In both cases, obstacles are assumed to be large clusters of events with distinctly different spatio-temporal features from those generated by ego-motion. Ball catching has been investigated using an event-driven visual input~\citep{wang2022ev}, combining a data-driven network and Kalman filter to predict the interception position for a 1D actuator with a net, but without POC analysis. Object tracking has been performed in cluttered environments with a moving camera using model-based Bayesian filtering, which avoids GPU latency~\citep{glover_robust}, while other methods bind latency for robot control tasks~\citep{glover_controlled}.
We built the event-driven perception pipeline exploiting speed invariance and limiting algorithm complexity~\citep{gava_toolate} for low-latency object tracking.

\subsubsection{Event-based perception} \label{sec:Event-based perception}
\begin{figure}
    \centering
    \includegraphics[width=.95\linewidth]{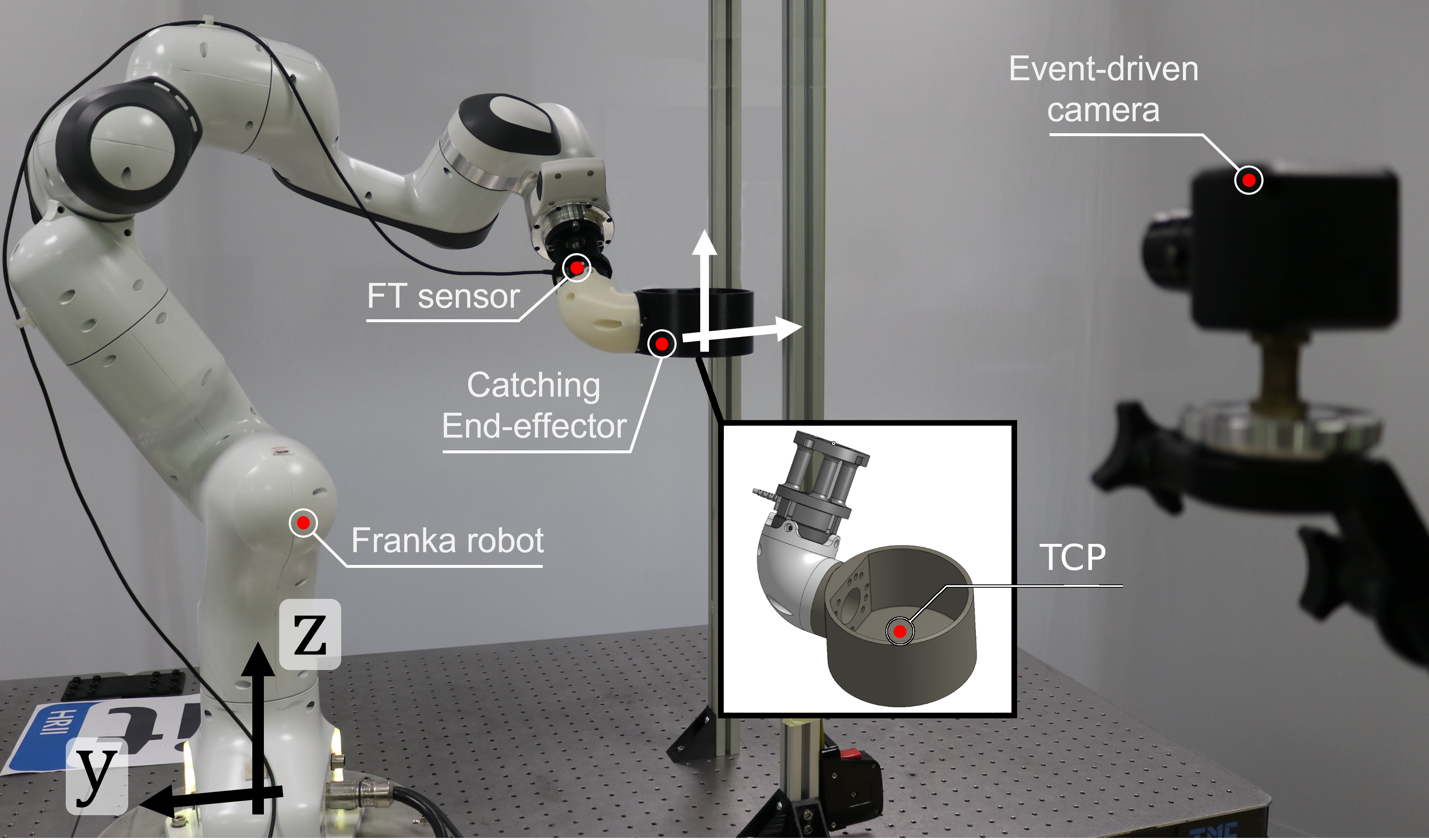}
    \caption{Experimental setup: the catching EE is connected to the robot through a force-torque sensor (ATI Mini45), with a combined mass of $0.47$ kg. The world z-axis is aligned with the z-axis of the EE, with origin in the Tool Center Point (TCP). The sensor was added since forces and torques estimated by the robot via motors' currents are inaccurate in impact studies. The event-driven camera is positioned and calibrated to be centered on the falling region.
    }
    \label{fig:1d-exp-setup}
\end{figure}

We use an ATIS Gen3 event-camera with a $640\times480$ pixels resolution, visible in Fig.~\ref{fig:1d-exp-setup}.
We use the parallel surface and convolution kernel (PUCK) algorithm~\citep{gava_toolate} to track the flying ball and generate object positions on the visual plane, $\langle u_t, v_t \rangle$ at time $t$.

The PUCK algorithm achieves low latency by reading \textit{events} $e = \langle u, v \rangle$ from the camera and updating an Exponentially Reduced Ordinal Surface (EROS) $EROS : (e_u, e_v) \mapsto [0.0, 1.0]$, with the size of the event-camera resolution. The EROS update is designed to minimize the computations performed \textit{per event} and be robust to variations in the object's velocity~\citep{gava_toolate}.

A likelihood $L$ of object position is generated for any given pixel in the EROS based on the dot product of the region in the EROS centered on pixel $\langle i, j \rangle$ and an object template, $K$. $L : (u, v) \mapsto EROS_{ij} \cdot K$, therefore, forms the convolution of EROS with the object template. Assuming the target cannot move far between updates (as the updates are very fast), the convolution can be performed only in a small (10 pixels) region of interest around $\langle u_{t-1}, v_{t-1} \rangle$. The position on the image plane is $\langle u_{t}, v_{t}\rangle = argmax(L)$. For tracking a ball, the object template is a circle of the approximate size of the ball projected on the image plane.

The EROS update and the convolution are performed in parallel on two separate threads, and the output of the tracked object on the image plane is above $500Hz$.

A transformation is needed to calculate the target position in the robot reference frame, $\langle u, v \rangle \mapsto \langle x, y, z \rangle$. Assuming the camera and the robot's operating spaces are restricted to 2D, a mapping from pixels to the robot's Cartesian coordinates is obtained through a visual calibration process. The calibration procedure acquired 30 pairs of values (pixel coordinates, robot position) in the task range. The mapping function is computed using a least squares optimization. The final $\langle x, y, z \rangle$ coordinates are sent to the trajectory estimator.

\subsubsection{Flying-object trajectory estimation} \label{sec:estimator}

The trajectory filtering and estimation are done by applying a linear Kalman Filter with the control input being gravity \citep{Gardner2022flyingObjectsEstimation}. Given, at instant $k$, the estimated state vector $\hat{\mathbf{x}}_k \in \mathbb{R}^{n_x}$, the observation vector $\hat{\mathbf{z}}_k \in \mathbb{R}^{n_z}$, and the input vector $\hat{\mathbf{u}}_k \in \mathbb{R}^{n_u}$, where $n_x$ is number of states, $n_z$, the number of observed states, and $n_u$, the number of control signals, we can define the dynamic model by
\begin{equation}\label{eq:kf-dynamic-model}
    \hat{\mathbf{x}}_{k} = \mathbf{F} \hat{\mathbf{x}}_{k-1} + \mathbf{G} \mathbf{u}_{k} + \mathbf{w}_{k-1},
\end{equation}
where $\mathbf{F} \in \mathbb{R}^{n \times n}$ is the state transition matrix, $\mathbf{G} \in \mathbb{R}^{n_x \times n_u}$ is the control matrix, and $\mathbf{w}_{k-1}$ is the process noise with Gaussian probability distribution $p(\mathbf{w}) \sim N(0,\mathbf{Q})$, where $\mathbf{Q} \in \mathbb{R}^{n_x \times n_x}$ is the process noise uncertainty. Then, the measurement equation is defined by \eqref{eq:kf-measurement-equation} with $\mathbf{H} \in \mathbb{R}^{n_z \times n_x}$,
\begin{equation}\label{eq:kf-measurement-equation}
    \mathbf{z}_{k} = \mathbf{H} \hat{\bm{x}}_k.
\end{equation}
The state vector for the single-axis experiments is $\hat{\mathbf{x}}_k = [p_z, \dot{p}_z]^T$, while from the camera only position is measured $\mathbf{z}_k= [p_z]$. The initial guess for the states is $\hat{\mathbf{x}}_0 = [0.7, 0]^T$, the input vector $\mathbf{u}_{k}=[g]$, and the matrices are given by:
\begin{equation}\label{eq:kf-matrices-1d}
    \begin{gathered}
        \mathbf{F} = \begin{bmatrix}
                        1 & dt \\
                        0 & 1
                    \end{bmatrix}, 
        \mathbf{G} = \begin{bmatrix}
                    0.5dt^2 \\
                    dt
                    \end{bmatrix}, 
        \mathbf{H} = \begin{bmatrix}
                      1 & 0  
                    \end{bmatrix}
    \end{gathered}
\end{equation}

For the multi-axis experiments, we have the state vector given by $\hat{\mathbf{x}}_k= [p_z, \dot{p}_z, p_y, \dot{p}_y]^T$, $\mathbf{z}_k= [p_z, p_y]^T$, with its initial guess $\hat{\mathbf{x}}_0 = [0.3, 1, -1.6, 2.5]^T$, and the observed states $\mathbf{z}_k= [p_z, p_y]^T$. Finally, the matrices are the following:
\begin{equation}\label{eq:kf-matrices-2d}
    \begin{gathered}
        \mathbf{F} = \begin{bmatrix}
                       1  & dt & 0 & 0 \\
                        0 & 1 & 0 & 0 \\
                        0 & 0 & 1 & dt \\
                        0 & 0 & 0 & 0
                    \end{bmatrix},
        \mathbf{G} = \begin{bmatrix}
                    0.5dt^2 \\
                    dt \\
                    0 \\
                    0
                    \end{bmatrix}, \\
        \mathbf{H} = \begin{bmatrix}
                      1 & 0 & 0 & 0 \\
                      0 & 0 & 1 & 0
                    \end{bmatrix}
    \end{gathered}
\end{equation}

\subsection{Robot parameters} \label{sec:exp:robotparams}
The experimental setup consists of a 7-DoFs Franka Emika Panda robotic arm with a 3D-printed flat EE, shown in Fig. \ref{fig:1d-exp-setup}. The tool's shape and its flat orientation at the catch, help to accentuate the catching impact along the vertical axis, to assess the proposed strategy's effectiveness. 
In our previous work \cite{zhao2022impact}, with a similar EE, we identified the wrist joints (joints 5 to 7) as the main limiting factors: their maximum torque is only $14\%$ of the first four joints \citep{FrankaDoc}. Particularly problematic was joint 6, which was highly loaded and often led to the robot's emergency lock. Therefore, in this work, the catching EE is designed with a 45$^o$ rotation (see Fig. \ref{fig:1d-exp-setup}) to provide a more uniform distribution of the internal torques across multiple joints.

The maximum payload at the EE declared by the robot manufacturer is $3~kg$, considering a static scenario \citep{FrankaDoc}. However, during fast movements, the additional inertial loads are nonnegligible, affecting the effective payload in the direction of the movement. This might lead to failure when the safety system is triggered due to excessive torque demands for the motors. The declared EE's maximum velocity of $1.7~m/s$ was never achieved in the constrained task space used in these experiments, even when requiring a highly reactive behavior by commanded high torques. Therefore, the limits on velocities and accelerations at the planning level were reduced to the empirically tested ones. In particular, for the velocities in \eqref{eq:plannerlimit}: $\dot{x}^{max}_{lin}=1.0~m/s$ and $ \dot{x}^{max}_{ang}=2.5~rad/s$, while for accelerations: $\ddot{x}^{max}_{lin}=6.0~m/s^2$ and $\ddot{x}^{max}_{ang}=25.0~rad/s^2$.

Since the learned human stiffness depends on the contact momentum~\citep{lacquaniti1989role,zago2009visuo}, a proper object weight should be chosen to transfer the learned stiffness to the robotic arm. Scaling to the robot's weight might result unreasonable, since this is affected by various aspects, such as the robot's manufacturer, materials, motor types, robot type, etc. Instead, we believe it is more reasonable to scale the object's weight with respect to the robot's payload since the payload is the most relevant capacity parameter of the robot, in this case. The catching object is a 3D-printed, hard-plastic ball with a mass of $0.1~kg$ and a diameter of $62~mm$. The weight of this ball is around 4\% of the actual payload $2.53~kg$ of the robot (the maximum payload of $3~kg$ minus the combined mass ($0.47~kg$) of the EE and force-torque sensor). In this way, and thanks to the considerations in Sec. 4.1, we achieved a one-to-one catching-object-weight-to-payload mapping from the human to the robot side.
    Moreover, the stiff contact between the tool and the object brings additional challenges since no passive dissipation mediates the impact.
    For this reason, the weight of the ball used in the experiments is designed to generate high peak forces, that exceed the robot limits even with relatively small releasing heights.
    This is useful to test the proposed strategy in the most adverse conditions and assess its effectiveness in reducing the impact forces, so that with a lighter/softer ball the performances can only improve.

In the POC phase, the parameters are set for the free-falling (multi-axis) case as $\Delta t_{poc}=0.2s \ (0.15s), d_{lim}= 0.13m \ (0.09m)$, and $K_d^{max}=750N/m, K_p^{max}=45, F_{th}=3N$ for both cases. The impact force is measured by the force-torque sensor mounted at the EE of the robot (Fig. \ref{fig:1d-exp-setup}) with a sampling frequency of 1k Hz. We assume that the impact occurs when the pre-defined threshold ($F_{th}$) is reached and that it can be detected by the force-torque sensor immediately. Based on the above parameters, the learned HVS was scaled by Algorithm \ref{alg:stiff}. Furthermore, to ensure smooth changes, the scaled stiffness was filtered as
\begin{equation}
    \bm{K}_P^{k}=\varepsilon\bm{K}_P^{k}+(1-\varepsilon)\bm{K}_P^{k-1}, \nonumber
\end{equation}
where $\bm{K}_P^{k}$ and $\bm{K}_P^{k-1}$ are the control gains at $k$ and $k-1$ instants, respectively, and $\varepsilon=0.05$ is the filtering parameter.

\subsection{Planner and Controller parameters}

The QP solver used for the online motion planner and the HQP controller is the open-source OSQP in C++, given its low computational times and sufficient accuracy. Both the controller and the planner run in real time at $1 ~kHz$. In particular, the planner's maximum calculation time always remains under $1~ms$, and its peak occurs at the beginning of the trajectory, when the estimated catching point is far (large $T_p$), and the optimization's size is larger. 
The trajectories generated have a lower sampling $\Delta t_p = 10~ms$ to maintain the real-time feasibility of problem \eqref{eq:general_QP_formulation}.
Given the short time window of the impact, the purpose is to provide fast updates based on the perception data. 
Hence, a middle layer is built to interpolate the trajectories sampled at $100~Hz$ into $1~kHz$ for the controller. 
When RMM is active, the number of layers in the controller's SoT is $N_p = 2$, where the first one is \eqref{eq:min_kyn_clik} and the second is \eqref{eq:min_mi_index}.
All experiments were performed in Ubuntu 20.04, with an Intel Core i7-11700 2.5 GHz $\times$ 16-core CPU and 32 GB RAM.

\begin{figure*}[t]
    \centering
    \includegraphics[width=\linewidth]{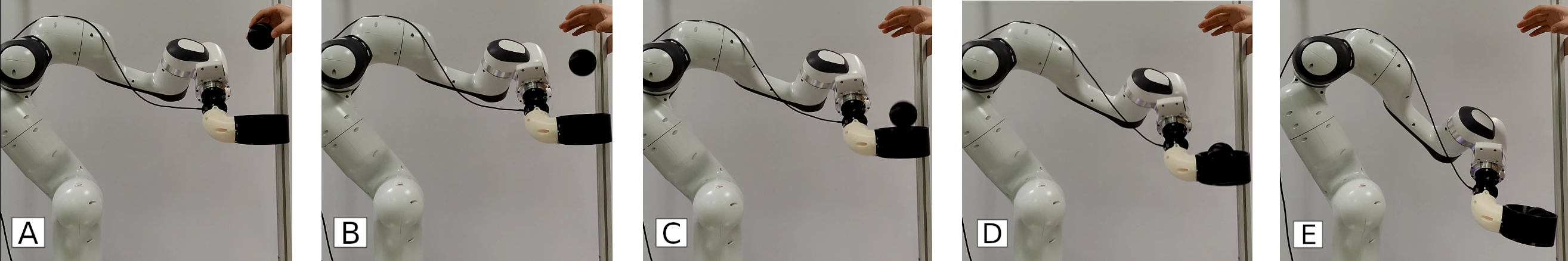}
    \caption{Time sequence of the VM-VIC one-dimensional catching task. PRC phase: the robot starts to move from its initial configuration (A) as the object is detected (B). The impact occurs in (D). POC phase: the robot follows the learned position, velocity, and impedance behaviors until the final point (E).
    }
    \label{fig:1d_snapshots}
\end{figure*}

\subsection{Metrics definition}
To accurately analyze the performances of the proposed method, we use the quantitative metrics proposed in \cite{zhao2022impact} in Cartesian space and shown in Table \ref{tab:1d-metrics-results}.

The `lift off index' (LOI) is the integral of the difference between the vertical component of the wrist force $F_z$ and its steady state value (which corresponds to the weight of the ball $F_w$) in the time interval $\Delta t_{poc}$, i.e., from the impact until steady state. A higher LOI indicates multiple bouncing and/or long under-damped catching behavior.
The `damping ratio index' (DRI) is the logarithmic decrement between the first and second impact force peaks $f_{z_{p,1}},f_{z_{p,2}}$, with $\delta=\log \big ( \frac{f_{z_{p,1}}}{f_{z_{p,2}}} \big )$, which indicates the capability to absorb impact force and energy and damp it quickly.
The `bouncing time index' (BTI) is the sum of all time intervals $\Delta \hat{t}$ in which the contact between the ball and the robot is lost in $\Delta t_{poc}$. $F_{max}$ is the maximum force, while the `velocity matching error' (VME) is the difference between the robot's $\dot{\bm{x}}^{c}_a$ and object's $\dot{\bm{x}}_{o}^{c}$ actual velocities at the effective impact.
Finally, $\Tilde{\bm{x}}$ is the difference between the actual robot's position at the effective contact $\bm{x}_a^c$ and the estimated object's position at the estimated catching instant $\bm{x}_o^{t_c}$.

\section{Experiments and Results}\label{sec:experiments}

This section presents the single- and multi-axis experiments. After introducing the notation, the comparative experiments with single-axis free-falling catching are analyzed. We select the best candidate and check its behavior by adding RMM and increasing the dropping height for robustness. Finally, the chosen candidate is tested in a multi-axis experiment, where the addition of RMM is further evaluated.

\subsubsection*{Experiments notation:}

this helps to identify the methods used in each experiment with a combination of PRC and POC as \textit{PRC-POC(-DIM)}, where DIM is the addition of RMM.
We summarize the strategies in the following lists:

\begin{itemize}
\setlength{\itemindent}{-.2cm}

    \item[] \hspace{-.7cm} \textbf{PRC}
    
    \item[-] \textbf{Fixed-position (FP)}: the planner outputs only a constant position estimated for the catching point (null velocity). 
    
    \item[-] \textbf{Velocity matching (VM)}: the planner outputs an optimal trajectory until the impact occurs, focusing on minimizing the relative robot-object velocity in \eqref{eq:impact_relationship}.
    
\end{itemize}

\begin{itemize}
\setlength{\itemindent}{-.2cm}
    \item[] \hspace{-.7cm} \textbf{POC}
    
    \item[-] \textbf{Impedance controller (IC)}: the QP-based impedance control defined in \eqref{eq:min_kyn_clik} is used with constant stiffness and damping terms. The terms
    K\textsubscript{H} and K\textsubscript{L} are used to indicate high and low constant stiffness values' choice.

    \item[-] \textbf{Switching Impedance controller (SIC)}: 
    this is a trivial variable impedance controller to track the trajectory as accurately as possible with high gain K\textsubscript{H} until the impact. After that, it switches to low stiffness K\textsubscript{L} to create a more compliant behavior and avoid bouncing.

    \item[-] \textbf{Variable impedance controller (VIC)}: uses the method proposed in Sec. \ref{sec:poc}, with state-dependent variable impedance learned from human demonstrations, together with planned position and velocity trajectories. In this case, the impedance is initially high K\textsubscript{H} to achieve accurate trajectory tracking in PRC, it then changes at the impact based on the learned human behavior, eventually returning to its original value.

\item[]  \hspace{-.7cm} \textbf{RMM}
    \item[-] \textbf{Dynamic Impact Measure (DIM)}: activation of the DIM maximization \eqref{eq:min_mi_index} to study its effect on both the exchanged forces \eqref{eq:impact_relationship} and the internal joint torques, as stated in Sec. \ref{sec:mrm}.
\end{itemize}

\subsection{Free-falling object} \label{sec:exp1D}

The first scenario we study in this section is limited to a single axis. The goal is to establish a repeatable experimental setup to thoroughly assess each proposed component's effect on the multiple performance metrics.

\begin{figure*}[t]
    \centering
    \includegraphics[width=\linewidth]{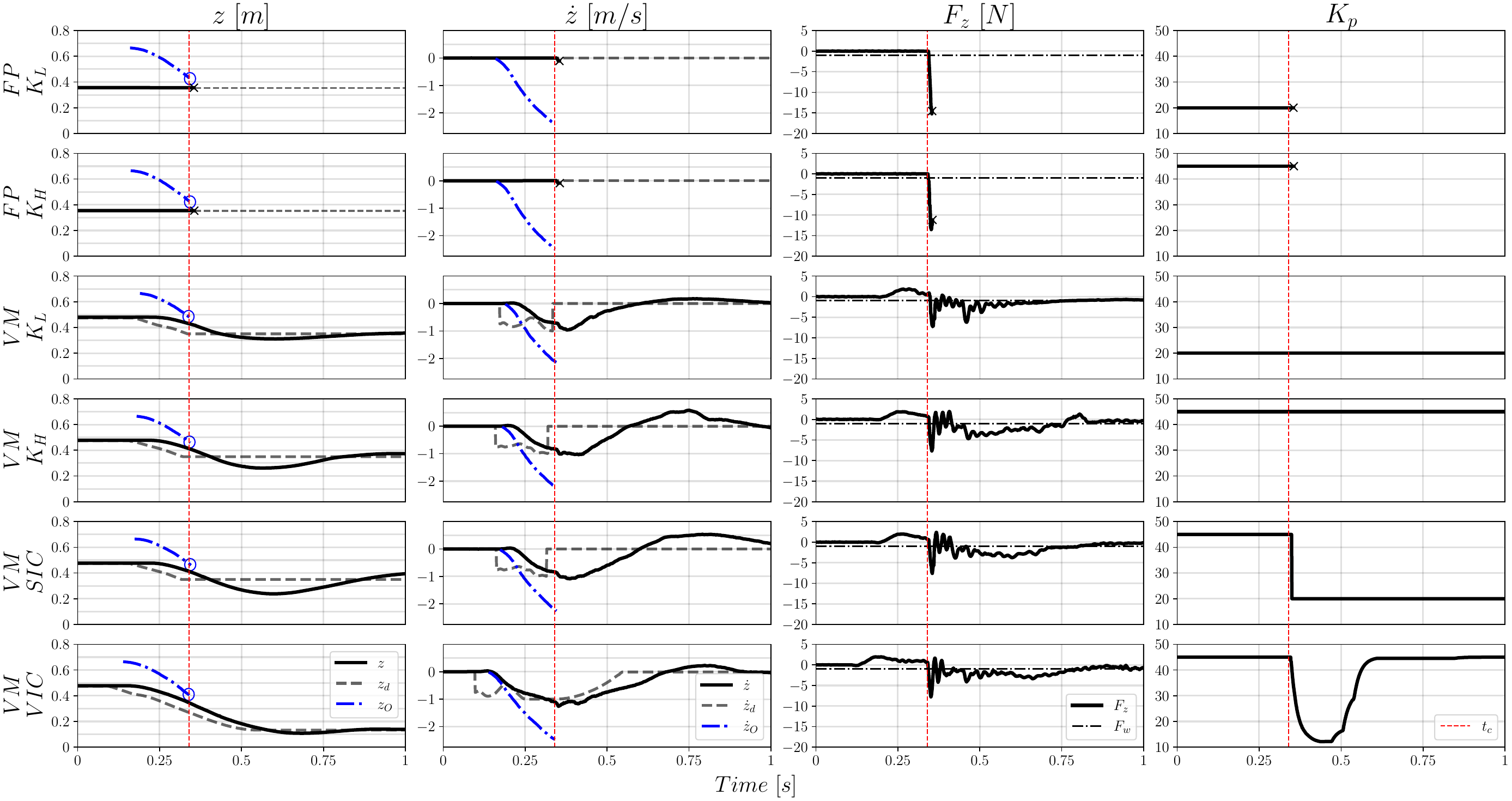}
    \caption{Comparison plots for the free-falling object experiment, synchronized by impact time: the first column contains actual ($z$) and desired ($z_d$) robot's position, and filtered object's position ($z_O$); the second column has the actual ($\dot{z}$) and desired ($\dot{z}_d$) robot velocities, and object's velocity ($\dot{z}_O$); the third column shows the interaction force ($F_z$) measured in the world frame and the object's weight ($W_O$); finally, fourth column shows the controller's stiffness ($K_p$) along $z$. The red vertical dotted line is the instant in which collision occurs.
    }
    \label{fig:1d_comparison_plots}
\end{figure*}

\begin{figure}
    \centering
    \includegraphics[width=\linewidth]{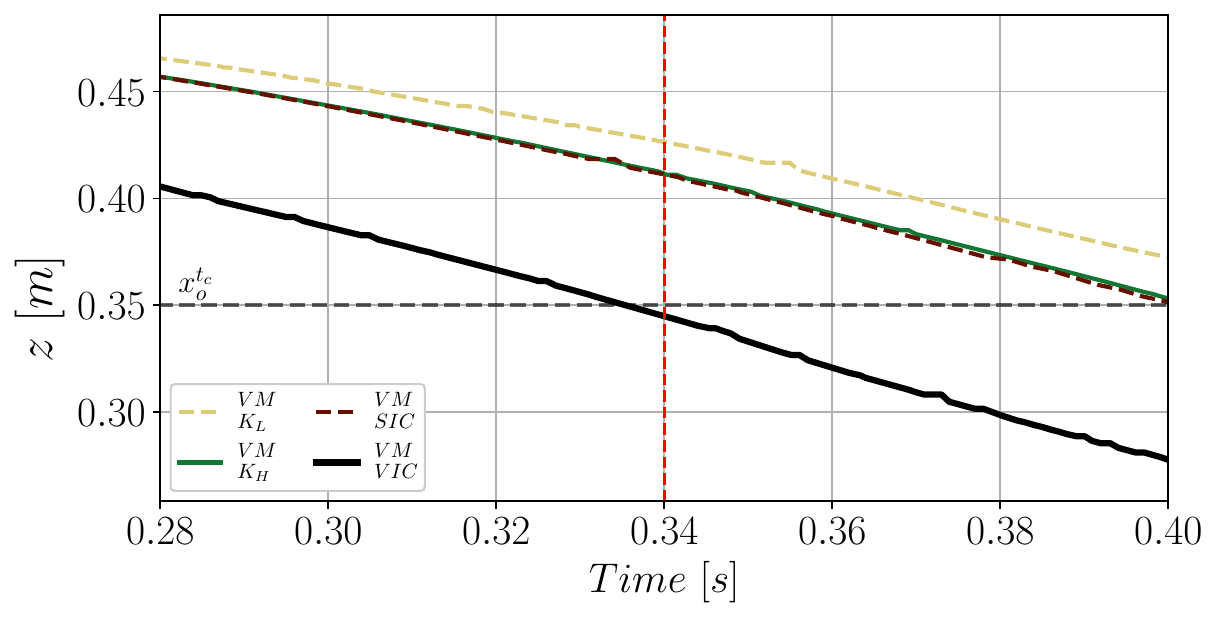}
    \caption{Zoomed plot around the collision moment (dashed red vertical line) of the robot's actual position for each controller with VM method in PRC. The horizontal line depicts the desired catching point $x_o^{t_c}=0.35~m$.
    }
    \label{fig:1d_zoom_position}
\end{figure}

\begin{figure}
    \centering
    \includegraphics[width=\linewidth]{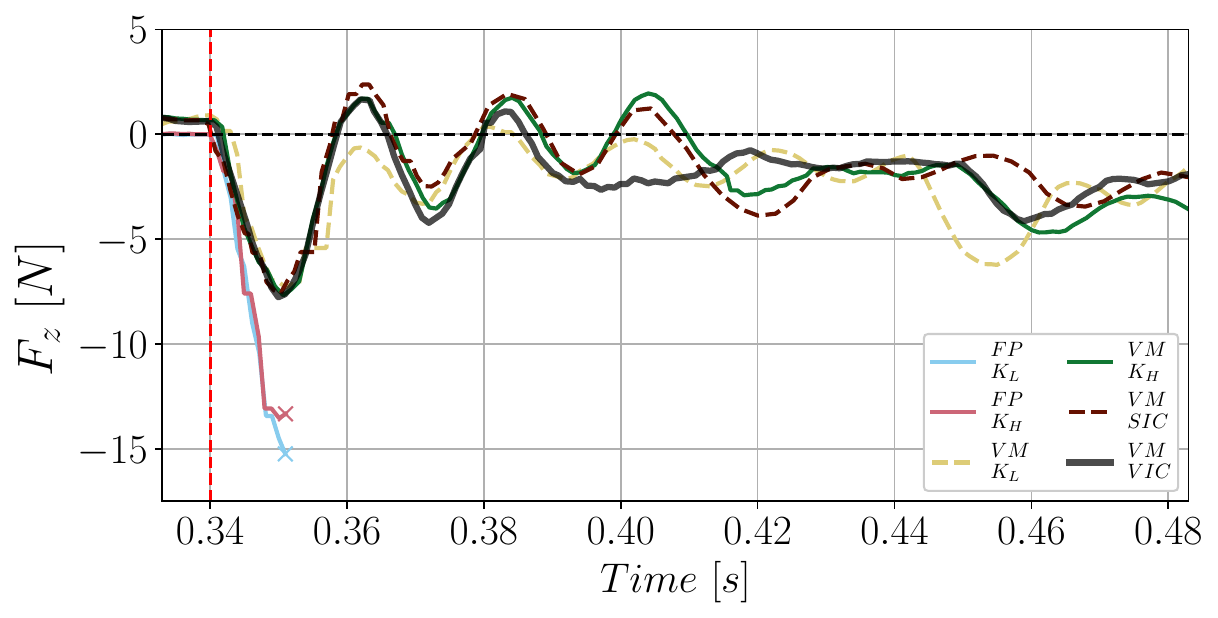}
    \caption{Detail of impact force for the six proposed methods. Apart from the FP experiments that stop at the impact, VM-K\textsubscript{L} is the only controller that does not lose contact at the expense of having a higher peak due to poor VM. Meanwhile, our proposal VM-VIC has a lower peak force and short periods without contact compared to VM-K\textsubscript{H} and VM-SIC.
    }
    \label{fig:1d_forces_impact}
\end{figure}

\renewcommand{\arraystretch}{1.}
\begin{table*}[ht] \caption{Metrics and results of the free-falling object experiments}
\centering
\begin{adjustbox}{} \label{tab:1d-metrics-results}
\begin{tabular}{lccccccc}
           &  \textbf{LOI} [Ns]               &  \textbf{DRI} [-]  &  \textbf{BTI} [ms]            &  \textbf{F}$_{max}$ [N] &  \textbf{VME} [m/s] & $\Tilde{\bm{x}}$ [m]\\ \hline
          \\ 
           &  $\int_{\Delta t_{poc}}|F_z-F_w| dt$    &   $(1+(\frac{2\pi}{\delta})^2)^{-0.5}$       &   $\sum_{j=1}^{n} \Delta \hat{t}_j$         &   $\max(|F_z|)$ & $| \dot{\bm{x}}^{c}_a - \dot{\bm{x}}_{o}^{c} |$   & $|\bm{x}_a^c - \bm{x}_o^{t_c}|$\\
          \\ \hline
\textbf{FP-K\textsubscript{L}} & -            & -             & -               & 15.2         & 2.48        & -     \\
\textbf{FP-K\textsubscript{H}} & -            & -             & -               & 13.5         & 2.54        & -     \\
\textbf{VM-K\textsubscript{L}} & \textbf{0.52}& 0.16          & \textbf{6}      & \textbf{7.3}          & 1.45        & 0.075 \\
\textbf{VM-K\textsubscript{H}} & 0.98         & 0.15          & 86              & 7.7          & 1.41        & 0.059  \\
\textbf{VM-SIC}                & 0.78         & \textbf{0.23} & 24              & 7.6          & 1.42        & 0.057  \\
\textbf{VM-VIC}                & 0.72        & 0.12          & 16              & 7.8          & $\bm{1.40}$ & $\bm{0.007}$ \\ \hline
\end{tabular}
\end{adjustbox}
\vspace{-2.5 mm}
\end{table*}

\subsubsection{Comparative experiments:} \label{sec:exp1D:ablation}

we evaluate six combinations among PRC and POC: FP-K\textsubscript{L}, FP-K\textsubscript{H}, VM-K\textsubscript{L}, VM-K\textsubscript{H}, VM-SIC, and VM-VIC (proposed framework). Fig. \ref{fig:1d_snapshots} shows the snapshots for the proposed framework, where the ball is released in (A), the impact occurs in (D), and the final position after POC is (E). 
The states recorded and shown in Fig. \ref{fig:1d_comparison_plots} from left to right represent the robot's and the ball's positions ($z$); the robot's and the ball's velocity ($\dot{z}$); the interaction forces ($F_z$) and the impedance control's stiffness ($K_p$). 
These experiments are synchronized by impact instant ($t_c$), depicted by the vertical dashed red lines.
The ball is dropped consistently from $0.67m$ in the world frame, with a relative distance of $20~cm$ above the robot, which starts at $0.47~m$ in the world frame.

\subsubsection*{Kinematics:} the first comparison is about the EE's position and velocity. The EE's actual and desired position ($z$ and $z_d$, respectively) are shown with the filtered object's position $z_O$ in Fig. \ref{fig:1d_comparison_plots} (first column), and a zoomed plot around the collision moment for every case is provided in Fig. \ref{fig:1d_zoom_position}. Both FP experiments are unsuccessful, regardless of their stiffness, since the safety brakes are triggered upon impact due to high joint torques generated by the force peak. Specifically for this configuration, joint number five reaches the safety levels \citep{zhao2022impact}.
When introducing the VM, however, the task is accomplished successfully. Still, it is noticeable how the low stiffness of VM-K\textsubscript{L} degrades the tracking quality since the EE error at the actual impact is $\Tilde{x} = 0.075~m$ (last column of Table \ref{tab:1d-metrics-results} and Fig. \ref{fig:1d_zoom_position}), given that the actual EE height is $0.425~m$ and the desired catching height is $0.350~m$ (specified as input of the estimator in Sec. \ref{sec:estimator} and passed to the planner in Sec. \ref{sec:methodology} via $\bm{x}_o^{t_c}$), chosen to match the FP-scenarios. Low-performance tracking would affect the success of the catching when extending to the multi-axis scenario.
VM-K\textsubscript{H} is more accurate at the contact, where the EE's height is $0.409~m$ (with an error of $\Tilde{x} = 0.059~m$), at the expense of a larger overshoot. This is mitigated in VM-SIC, where both accuracy until $t_c$ (EE's height of $0.407~m$ and $\Tilde{x} = 0.057~m$) and low overshoot are achieved. The complete proposed method is shown by VM-VIC, where the POC phase starts as soon as the estimated trajectory is finished, with the stiffness profile learned from the human and properly adjusted. If an impact is detected beforehand, the POC phase is triggered accordingly.
The EE's height for the VM-VIC at the collision instant is $0.343~m$, and the position error is $\Tilde{x} = 0.007~m$,
corresponding to a reduction of $86\%$ when compared to the previous cases.
This difference with respect to VM-K\textsubscript{H} and VM-SIC cases, which are the other scenarios with the same stiffness in PRC, is related to the different position/velocity trajectories in the timespan ranging from the estimated to the actual impact. Indeed, in VM-K\textsubscript{H} and VM-SIC, once the estimated trajectory is finished, the previous position and zero velocity (for safety reasons) references are provided, until the impact is detected. With the proposed approach instead, the POC phase is triggered automatically, and the continuous velocity reference in the aforementioned timespan allows the robot to travel more than $5~cm$ lower, as visible from Fig.~\ref{fig:1d_zoom_position}.
This is further noticeable from the longer dropping time until the detected impact, in the position and velocity plots of Fig.~\ref{fig:1d_comparison_plots}.

The velocity analysis is based on the VME. The vertical components of the actual $\dot{z}$ and desired $\dot{z}_d$ robot's velocities are shown, together with the object's velocity $\dot{z}_O$ in the second column of Fig. \ref{fig:1d_comparison_plots}. The VME reported in Table \ref{tab:1d-metrics-results} shows that we achieve a reduction compared to the FP case of more than $1~m/s$, lowering the impact forces considerably. The values achieved among the different PRC strategies are mostly similar and consistent, which is reasonable since VME does not depend on the POC controllers. However, VM-K\textsubscript{L} has worse velocity tracking, resulting in a slightly higher VME, which is again detrimental for higher dimensional experiments.

It must be noted, however, that even with the more conservative limit values on $\dot{\bm{x}}^{max}$ and $\ddot{\bm{x}}^{max}$ (as mentioned in Sec. \ref{sec:exp:robotparams}) specified in the optimal planner, the robot is still not capable of reaching such limits, especially in terms of acceleration, which is the main cause of the delay in the velocity tracking, and, consequently, of the error between the estimated and actual catching position.

\subsubsection*{Forces:} we now compare the forces under the presented metrics to evaluate each POC method performance. Firstly, both FP trials have the highest registered peaks with similar maxima by considering the forces measured along the $z-$axis (see Fig. \ref{fig:1d_forces_impact}). This was expected since the EE's stiffness does not affect the force peaks during the very short time window of the impact, verified by the small variations in $F_{max}$ between FP-K\textsubscript{L} and FP-K\textsubscript{H}, and between VM-K\textsubscript{L} and VM-K\textsubscript{H}.
Meanwhile, the addition of velocity matching during PRC, led to a strong reduction of approximately $50\%$ in terms of F$_{max}$ in all VM methods compared to FP.

The LOI metric indicates the damping properties in the timespan ranging from the impact to steady state. A lower LOI implies improved damping and shorter settling periods. VM-K\textsubscript{L} presents, as expected, the smallest metric ($0.52$), being very compliant, as opposed to VM-K\textsubscript{H} ($0.98$) and VM-SIC ($0.78$), with an increase of $88\%$ and $50\%$ respectively. Meanwhile, the proposed VM-VIC ($0.72$) increased only of $38\%$. The detail of the interaction forces after impact in Fig. \ref{fig:1d_forces_impact} shows several oscillations for VM-K\textsubscript{H} and VM-SIC (and higher LOI), followed by VM-VIC and VM-K\textsubscript{L}, respectively.

Moreover, LOI is affected by the robot's position and velocity tracking during POC: due to the nonprehensile catch, the object moves freely and affects the interaction forces, increasing LOI. This can be observed in VM-K\textsubscript{H} and VM-SIC with the large overshoots in position ($t \approx 0.6~s$) and velocity ($t \approx 0.75~s$) in Fig. \ref{fig:1d_comparison_plots}. VM-VIC's better tracking during POC also decreased the overshoots, indicating a damped transient until the steady state, with a better nonprehensile catch.

DRI indicates the capability to dissipate the impact force quickly and, therefore, should be maximized. The force signal shows a highly dynamic response since the time elapsed between the first and second peaks is very short, in the order of 30 ms (see Fig. \ref{fig:1d_forces_impact}). The quick change in stiffness of VM-SIC allowed it to achieve the highest DRI of 0.23, moderately higher than the other POC methods: VM-K\textsubscript{L} had $0.16$; VM-K\textsubscript{H}, $0.15$; and VM-VIC, $0.12$.
Although VM-SIC registered the highest DRI (fastest dissipation) this is at the expenses of a discontinuous stiffness variation, and its overall behavior is not the most damped, as indicated by the other metrics.
Besides, while we confirmed above how the robot's stiffness does not immediately affect the impact force \citep{haddadin2009requirements}, it still remains unclear when and how this compliance comes into play after the impact. The DRI could give a further insight in this sense, being related to a short time window following the impact, however more focused studies are necessary on this subject.

BTI is the last metric, and it should be minimized since it relates to bouncing. Therefore, a more damped controller should lead to less bouncing and smaller BTI. 
VM-K\textsubscript{L} had the smallest bouncing interval ($6~ms$), indicating the importance of a compliant behavior in POC. At the same time, VM-K\textsubscript{H} ($86~ms$) and VM-SIC ($24~ms$) presented higher bouncing time, approximately 14 and 4 times higher than VM-K\textsubscript{L}, respectively. VM-VIC ($16~ms$) is only less than three times higher, as shown in Fig. \ref{fig:1d_forces_impact}.

Therefore, the best overall trade-off between these metrics is obtained by VM-VIC, which ensures catching kinematic accuracy by being among the best PRC metrics (VME, $\Tilde{\bm{x}}$) while reducing force metrics in POC (LOI, DRI, BTI, and $F_{max}$), thus lowering bouncing by regulating the stiffness and the displacement during the POC phase. By considering both the PRC and POC and seeking to optimize the analyzed metrics on both phases, VM-VIC is considered the best impact-aware framework evaluated. 
Therefore, we will evaluate its performance within additional components and scenarios, e.g., different heights and the addition of DIM.

\begin{figure}[t]
    \centering
    \includegraphics[width=\linewidth]{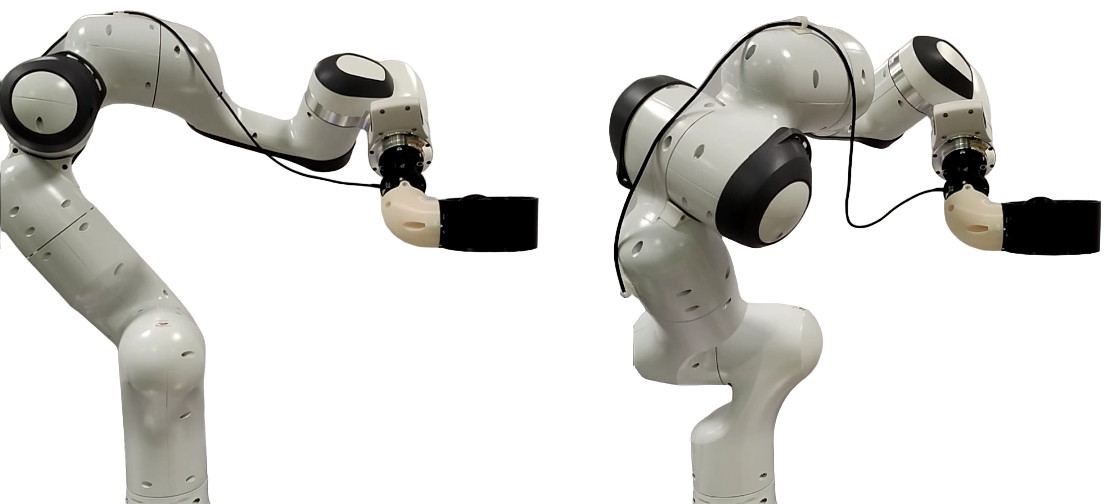}
    \caption{Configuration difference of the robot, considering equal EE position, for the cases without (left) and with (right) DIM optimization, respectively, corresponding to an increase in $w_{f_{d}}$ of $0.04$ (from left to right).
    }
    \label{fig:1d-exp-setup-rmm}
\end{figure}

\begin{figure}[t]
    \centering
    \includegraphics[width=\linewidth]{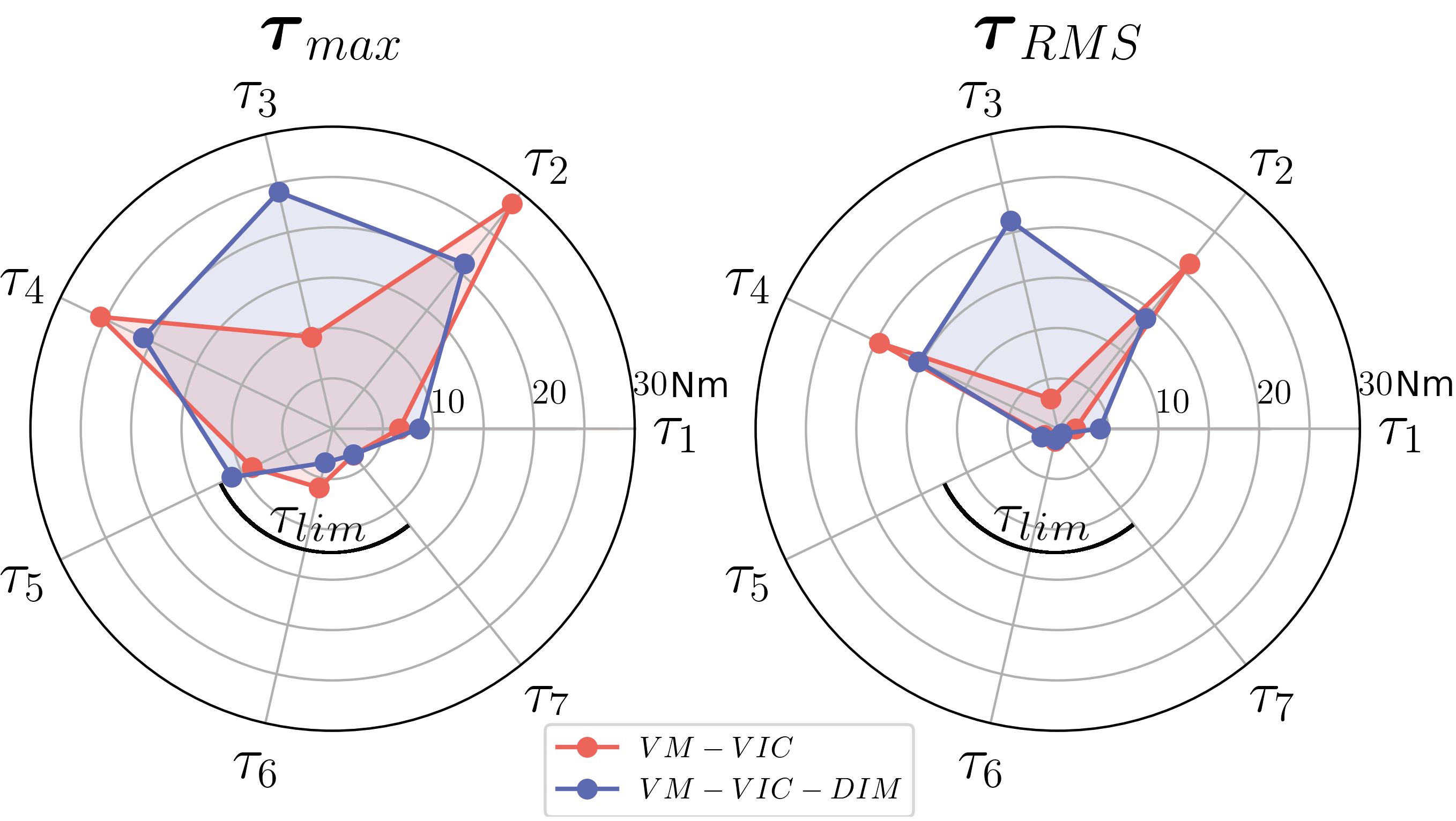}
    \caption{Polar plots comparing the maximum torques ${\boldsymbol{\tau}}_{max}$ (left) and the RMS torques ${\boldsymbol{\tau}}_{RMS}$ (right) for each joint. The torque limits of joints 1-4 is ${{\tau}}_{lim} = 87~Nm$, whereas for joints 5-7 ${{\tau}}_{lim} = 12~Nm$ (bold line).
    }
    \label{fig:1d_dim_polar_plots}
\end{figure}

\begin{figure}[t]
    \centering
    \includegraphics[width=0.95\linewidth]{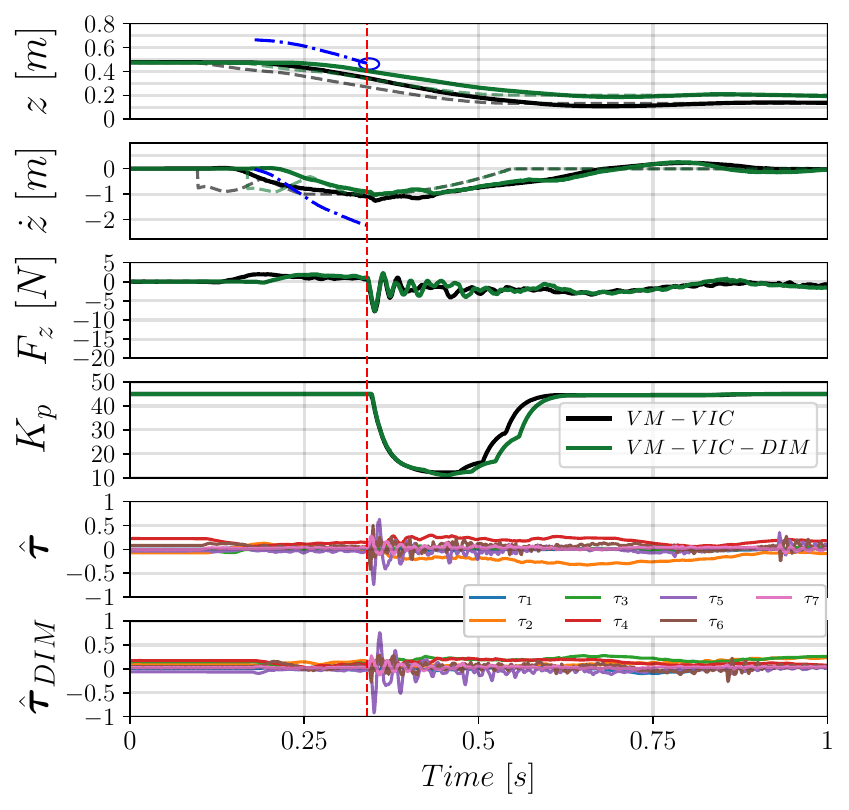}
    \caption{Experimental results for the activation of DIM maximization, depicted with collision time instant (vertical dotted red line). Besides the same signals presented in Fig. \ref{fig:1d_comparison_plots}, a final plot is added, showing the normalized actuation torques $\hat{\bm{\tau}}$. This experiment registered a peak force of $F_z=-7.7~N$ and VME $=1.36~m/s$.
    }
    \label{fig:1d_dim}
\end{figure}

\subsubsection{Influence of DIM optimization:} \label{sec:exp1D:rmm}

we add to the original QP problem \eqref{eq:min_kyn_clik} a secondary objective in the SoT to minimize the reflected mass in the impact direction \eqref{eq:min_mi_index}, identified by the direction of the estimated object's velocity at the catching point.
Therefore, the robot reconfigures online, going from the initial configuration of Fig. \ref{fig:1d-exp-setup-rmm} (left), to the one that optimizes the DIM, Fig. \ref{fig:1d-exp-setup-rmm} (right).
Fig. \ref{fig:1d_dim} shows the behavior of the states with the activation of the DIM, and Table \ref{tab:1dDIM-metrics-results} presents the metrics related to joint efforts.
The reduction in terms of peak forces measured at the EE is limited, going from $F_{max}=7.8~N$ to $F_{max}=7.7~N$ with the activation of DIM optimization. This is related to multiple aspects.
In particular, the small dropping height and extremely short time window of the task do not allow the robot to fully reconfigure. Besides, the forces generated at the EE are low (yet creating critical loading at the joints), leaving small possibility for improvement.
Therefore, we expect this benefits to increase even further when extending to the multi-axis scenario.
Besides, the dynamic impact ellipsoid varies inside the task space based on the robot's configuration, hence there are regions that will lead to higher/lower improvements in terms of RMM, given the different geometry of the ellipsoid.
Fig. \ref{fig:1d_dim_polar_plots} shows the polar diagrams for the cases with and without DIM activation, for both maximum ${\bm{\tau}}_{max}$ (left) and root mean square (RMS) ${\bm{\tau}}_{RMS}$ (right) torques. It is important to note that the motors of the Panda manipulator are not all the same. Indeed, joints 1-4 use bigger motors, with a torque limit ${{\tau}}_{lim} = 87~Nm$, whereas for joints 5-7 ${{\tau}}_{lim} = 12~Nm$, which is less than $14\%$ hence the smaller values of Fig.~\ref{fig:1d_dim_polar_plots}. Considering ${\bm{\tau}}_{max}$, we notice that the DIM activation allows a reduction of the torque peaks registered in most joints, especially joints 2, 4, and 6, that have their torques decreased by 27\%, 18\%, and 42\% (which is the biggest reduction, but less visible given the smaller ${{\tau}}_{lim}$), respectively, thanks to the activation of joint 3. This translates into a better torque redistribution among the actuators.
Although joint 3 increased by a factor of almost three, from $9.1~Nm$ to $24.1~Nm$, it is still working below 50\% capacity given that the joint's torque limit is $87~Nm$.

This is confirmed by ${\bm{\tau}}_{RMS}$, where the area covered by the polar plot provides a visual cue of the overall joints' activation. Comparing the two strategies, the activation of the most loaded joints throughout the task (i.e., joints 2 and 4, which indeed act against gravity) is reduced by 36\% and 21\%, respectively, thanks to the activation of joint 3, which was hardly active before. This results in a more uniform torque activation among joints and reduced stress on the most loaded motors.
The overall torque profiles are shown in the bottom plot of Fig. \ref{fig:1d_dim}, from which it is visible how, besides the first peak, the overall joints' RMS activation remains under $50\%$ of their respective limit, avoiding high and unbalanced activations of some motors.

To compare $w_{f_{d}}$ for each case, the `average DIM' (ADIM) is obtained by integrating it in the POC phase until steady state $\Delta t$ and taking its average over the experiment.
As noticeable from Table \ref{tab:1dDIM-metrics-results}, the ADIM is increased from $2.23$ to $2.27$ for the VM-VIC case. Despite the small variation, to provide a normalized value, it is necessary to map the DIM in the entire task space \citep{patel2015manipulator} since this is robot-dependent and unbounded. For simplicity, and lying outside the scope of our work, we only provide the relative values for each configuration since the actual difference in redundancy reconfiguration is substantial even for small relative values, as shown in Fig. \ref{fig:1d-exp-setup-rmm}.

Additionally, Table \ref{tab:1dDIM-metrics-results} provides a comprehensive overview of maximum and RMS torques across the joints. The results indicate a reduction of 16\% in the maximum absolute torque, from 28.6 Nm to 24.1 Nm. Simultaneously, the introduction of DIM leads to an increase in RMS torques across all joints by 19\%, particularly enhancing efforts in joints with higher torque limits. 

\renewcommand{\arraystretch}{1.1}
\begin{table}[t] \caption{Metrics and results with the addition of optimal DIM}
\centering
\begin{adjustbox}{width=.49\textwidth} \label{tab:1dDIM-metrics-results}
\begin{tabular}{lccc}
          & \large \textbf{ADIM} [-] & \large $\bm{{\tau}}_{max} $ [Nm] & \large $\bm{{\tau}}_{RMS}$ [Nm]  \\ \hline
          \\ 
          & $\frac{1}{\Delta t} \int_{\Delta t} w_{f_{d}}(\bm{q}) dt$    &  $\underset{n}{\max} ( \underset{\Delta t}{\max} (|\boldsymbol{{\tau}}|)$      &  $\sum_{i=1}^{7}\sqrt{\frac{1}{\Delta t} \int_{\Delta t} {\boldsymbol{\tau}}_i^2\, dt}$        \\
          \\ \hline
          \\
\textbf{VM-VIC}                     & 2.23        & 28.6         & 48.9             \\          \\
\textbf{VM-VIC-DIM}                 & 2.27   & 24.1  & 58.2     \\
\hline
\end{tabular}
\end{adjustbox}
\end{table}

\begin{figure} [t]
    \centering
    \includegraphics[width=0.95\linewidth]{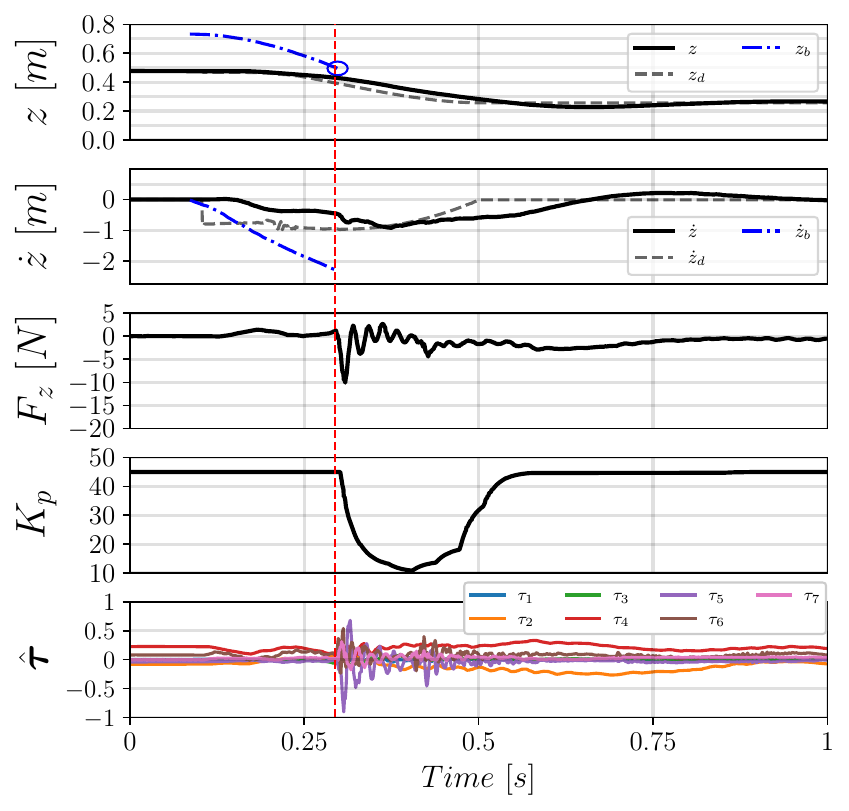}
    \caption{Experimental results for VM-VIC with increased height. Peak force $F_z=-10.0~N$ and VME$=1.90~m/s$
    }
    \label{fig:1d_increased_height}
\end{figure}

\subsubsection{Increased dropping height:} \label{sec:exp1D:increased_height}
 
to assess the method's behavior (VM-VIC case only) with a different setup, we increase the height from which the ball is dropped, as close as possible to the point in which the experiment is no longer feasible due to excessive impact forces, resulting in the safety locking of the robot. In this sense, a maximum increase of $10~cm$ in height is feasible, and we analyze its performance in Fig. \ref{fig:1d_increased_height} for the VM-VIC case. The results are similar to those already obtained, despite the object's kinetic energy increasing of 85\% given this is proportional to the square of the velocity, which led to a VME increase of 35\% ($1.90~m/s$). The new force peak is $10.0~N$ (28\% increase), demonstrating the beneficial PRC effect on the task, even for increased dropping height. The ball is caught $47~ms$ earlier than the estimated impact time, which also shows the robustness against the object's estimation.  Moreover, POC's importance is proven by the BTI increase to $50~ms$ and LOI to $0.65~Ns$ due to the variable stiffness, close to the previous VM-VIC. Finally, the normalized torques $\hat{\bm{\tau}}$ show a first impact peak close to the actuator's limit for joint 5, which was already identified among the most loaded ones in the previous experiments.

In conclusion, the proposed strategy managed to successfully catch up to a dropping height of $0.77~cm$, as opposed to the cases without VM. Thanks to the POC human-learned  behavior, it was possible to outperform other VM-based approaches by reducing bouncing and the reflected EE mass, while maintaining high position and velocity tracking accuracy.

\subsection{Dimensionality extension and generalization} \label{sec:exp3D}

After analyzing the results obtained from the single-axis scenario, the goal is to generalize the proposed framework to multiple dimensions to evaluate its behavior in space and further test its robustness. This setup requires high tracking capabilities to guarantee the catch success, therefore a high-speed perception system is necessary. The event-driven high frequency camera was important to feed data to the Kalman filter at a high update rate, allowing the planner to re-plan in real-time.
However, the absence of in-depth perception does not allow for obtaining complete three-dimensional information about the flying object. Therefore, to avoid stereo-cameras scenarios, which would noticeably increase the setup costs and lie outside the scope of this work, we choose to analyze the problem in the 2-D plane perceived by the camera. This is crucial to highlight the validity of the proposed method and provide proof of its generalization to multiple axes.

This new setup is more complex due to the random initial conditions of the object, as the human cannot guarantee repeatability. Therefore, it must be noted that, without the one-dimensional analysis, it would not have been possible to evaluate the effects of the proposed framework on the catching forces and robot's torques, due to this variability in the throwing, which randomizes the final catching point and impact forces. This experiment shows the successful multi-dimensional catching with VM and DIM's influence. However, given the aforementioned variability, no direct comparison between catching forces and robot torques can be drawn among the trials.
Hence, we employ the best techniques identified in Sec. \ref{sec:exp1D}, i.e., VM-VIC and VM-VIC-DIM, to validate these along multiple axes.

\begin{figure*}[t]
    \centering
    \includegraphics[width=1\linewidth]{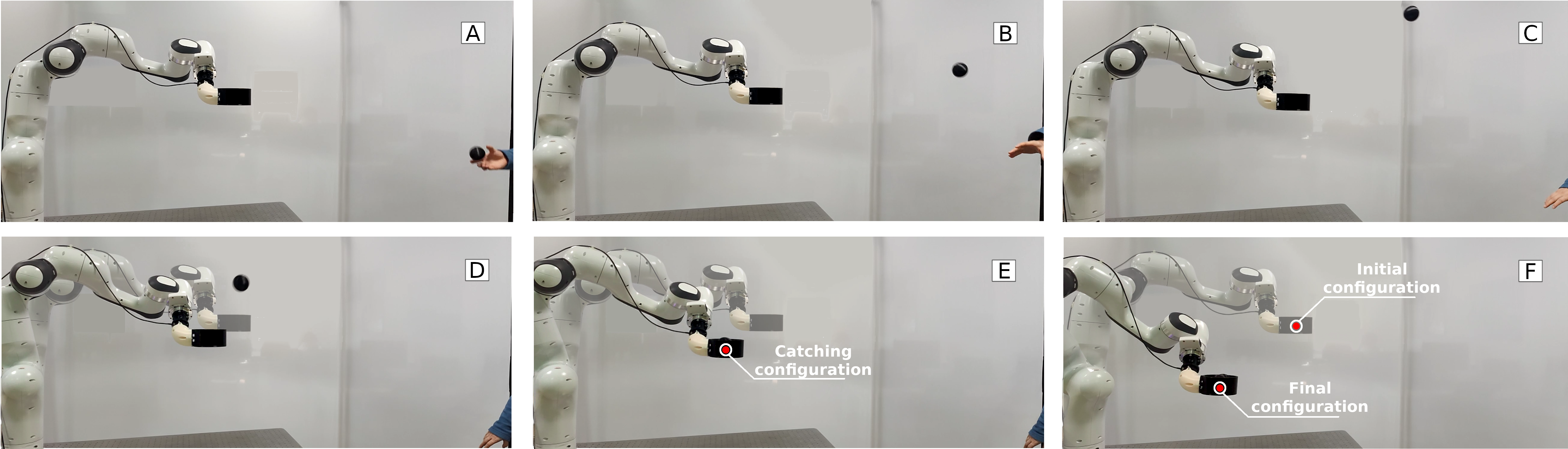}
\caption{Extension to multi-axis object-catching. The robot starts in the initial configuration (A). As the object is detected and the optimal plan to achieve VM is generated, the robot moves (C) up to the catching instant (E). After the impact, the POC phase is activated (F).
}
    \label{fig:2dsnaps}
\end{figure*}

\begin{figure*}[!ht]
    \centering
    \includegraphics[width=1\linewidth]{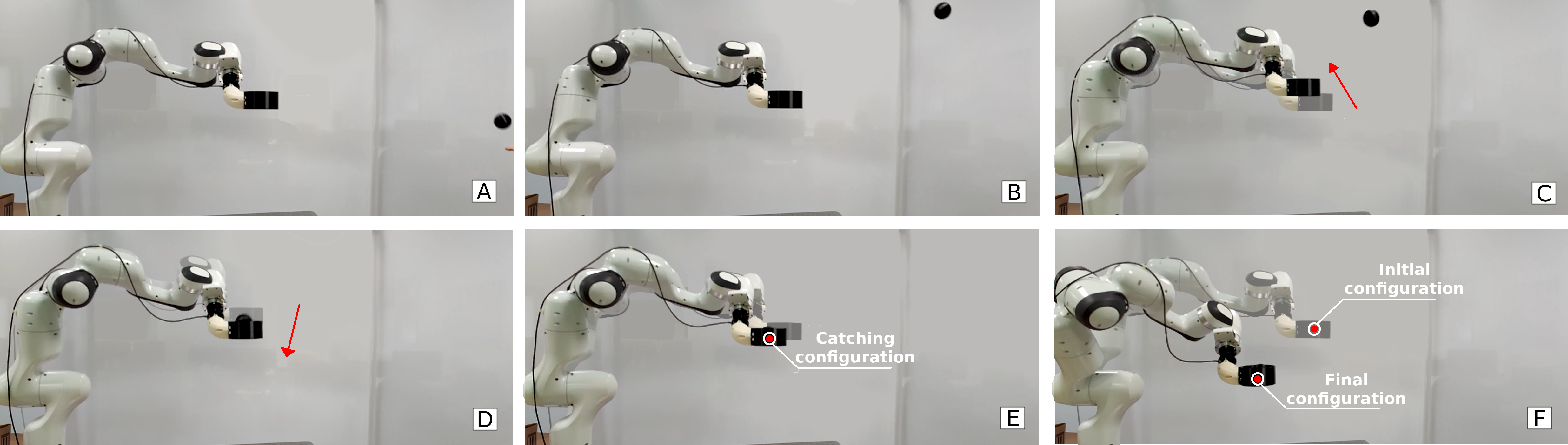}
\caption{Extension to multi-axis object-catching, with the optimal DIM active as a secondary priority of the HQP controller. In this example, being the final catching position already close to the initial one of the EE, to optimize VM, the robot starts moving upwards (C), then inverting its speed (D) to reach the catching point (E) with optimal VM. POC is triggered upon impact detection (E) to dampen the impact energy until (F). 
}
    \label{fig:2dRMMsnaps}
\end{figure*}

\begin{figure*}
    \centering
    \includegraphics[width=\linewidth]{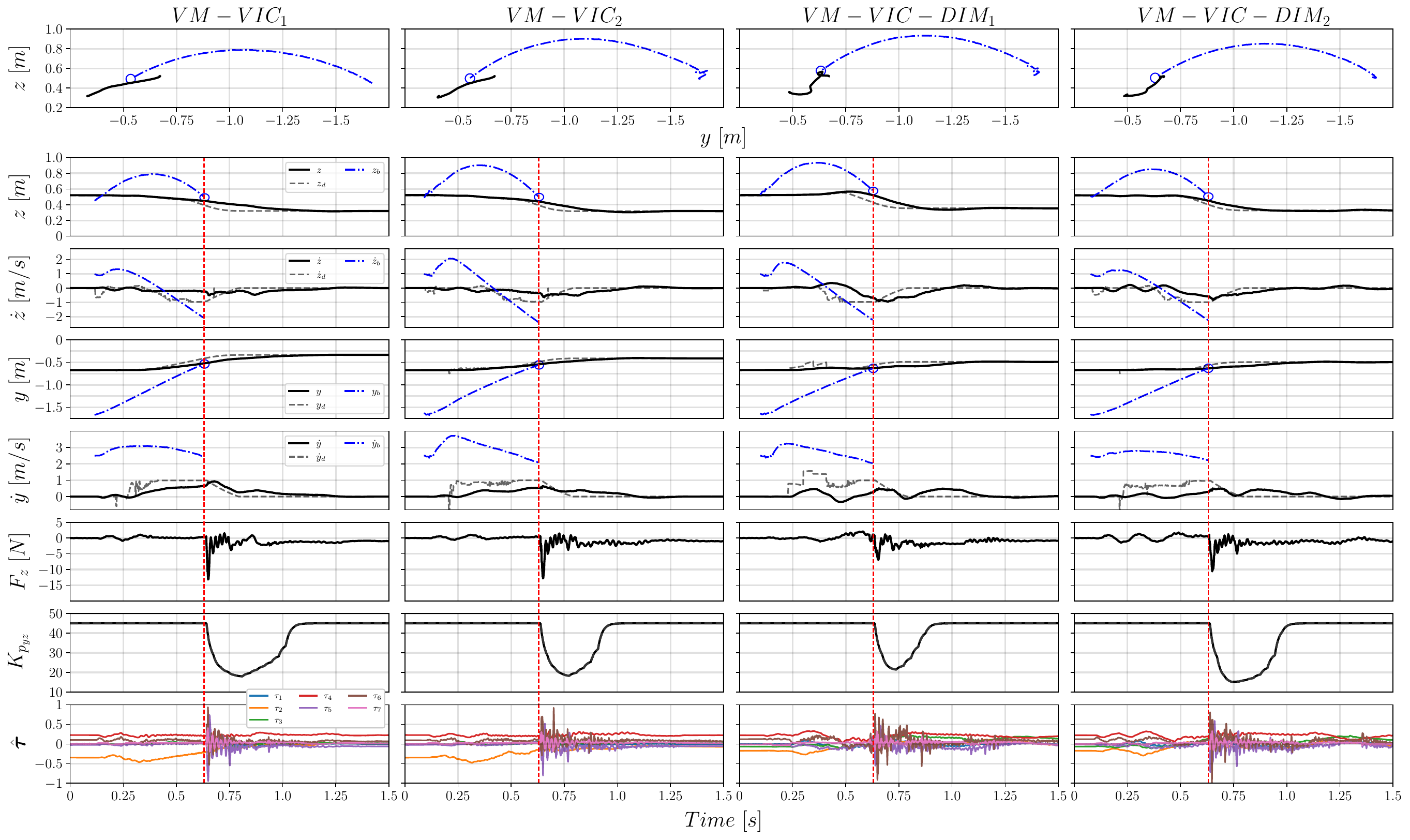}
    \caption{
    Comparison plots for the dimensionality extension experiments, synchronized by collision time instant $t_c$ (red vertical dotted line).
    Each column shows the results obtained for different trials: two trials for VM-VIC and two for VM-VIC-DIM.
    The first row shows the ball (dashed blue line) and EE (solid black line) trajectories in the Cartesian plane under consideration.
    The second row contains the actual ($z$) and desired ($z_d$) robot's height and the filtered object's height ($z_O$). The third row shows the actual ($\dot{z}$) and desired ($\dot{z}_d$) robot velocities and estimated object's velocity ($\dot{z}_O$). The same is shown in the fourth and fifth rows along the $y$-axis.
    The sixth row shows the interaction force ($F_z$) measured in the world frame. The seventh row shows the controller's stiffness in both directions ($K_{p_{yz}}$), and the final row shows the normalized actuation torques ($\hat{\bm{\tau}}$) in time.
    }
    \label{fig:2d_time_plots_v2}
\end{figure*}

\subsubsection*{VM-VIC:} Fig. \ref{fig:2dsnaps} shows the snapshots of the overall catching task for VM-VIC, from the PRC phase in which the ball is thrown (A-B-C-D) until the catching instant (E) and the POC phase (F) in which the difference between the initial and final configurations is shown.
The VM-VIC's results corresponding to the two trials of the sequence shown in Fig. \ref{fig:2dsnaps} are reported in the first two columns of Fig. \ref{fig:2d_time_plots_v2}.

Adding a new dimension and the randomness in the throw reflected directly on the metrics. Indeed, for both trials, the catching instant delay is $32~ms$ and $63~ms$, while the position error in the $yz$-plane at the impact instant is $0.130~m$ and $0.077~m$.
The ball's dropping heights are $0.84~m$ and $0.95~m$, as opposed to the fixed height of $0.67~m$ of the free-falling 1DoF experiment. In the $y$-axis instead, the ball traveled $1.12~m$ and $1.09~m$, respectively. This increased distance results in the increasing catching velocity of the ball and, consequently, increasing VME in both cases to $2.48~m/s$ and $2.62~m/s$. This leads to higher interaction forces $F_{z}$ of $13.2~N$ and $12.9~N$. Although the force peak is higher than the limit obtained in the increased height experiment (Sec. \ref{sec:exp1D:increased_height}), new joints are activated in this motion, changing the internal peak torques distribution during the movement, as shown in the last row of Fig. \ref{fig:2d_time_plots_v2}. Apart from joint five, which was highly activated during the one DoF experiment, joint 6 raised its contribution, allowing a higher impact force.

\subsubsection*{VM-VIC-DIM:} the same experiment is repeated twice, activating optimal DIM. The snapshots are reported in Fig. \ref{fig:2dRMMsnaps}, and we show, in this case, a trial in which the estimated catching point is already close to the initial EE position (A), as visible from (E). The aim is to highlight the VM behavior since, instead of simply remaining in the catching position, the robot moves in the opposite direction (C) and eventually moves back (D) to improve the VM.
The results are shown in the last two columns of Fig. \ref{fig:2d_time_plots_v2}, with errors in the catching instant delay of $32~ms$ and $67~ms$, while the position error in the $yz$-plane at the impact instant is $0.118~m$ and $0.089~m$. These experiments had the ball reaching $0.98~m$ and $0.90~m$ in $z$, and traveling $1.02~m$ and $1.03~m$ in $y$. Meanwhile, the resulting VME is $2.37~m/s$ and $2.56~m/s$. VM-VIC-DIM reduced the catching force in both trials: $6.9~N$ and $10.6~N$.

As already discussed, an horizontal reference orientation is provided to the EE. 
However, the planning problem remains valid for any other orientation, thanks to the definition of the generic desired catching pose $\bm{x}_o^{t_c}$ in \eqref{eq:general_QP_formulation}.
The same can be said for the controller, in which the RMM can occur for any impact orientation, which is specified via the versor $\bm{u}$ in \eqref{eq:dm_index}. In this case, we maintain this direction purely vertical as before, only to isolate its effect along the vertical axis and operate a proper comparison with the previous experiments.

\section{Conclusion}\label{sec:con}
This work focused on the problem of nonprehensile object catching and introduced an impact-aware control for pre- and post-catching phases.  
Current nonprehensile applications range from catching/throwing to batting and manipulating objects.
The potential applicability is wide, for example, in large industrial plants, where objects are transported via conveyors and need to be sorted/separated on the fly, or in garbage separation for recycling. 

Using our proposed framework, the interaction forces were minimized by starting from the PRC phase, through the VM between the object and the end-effector. Optimal and real-time motion planning was necessary, given the short time window in which the object was in flight. This increased the catching accuracy as the object's state estimation improved with event-driven cameras.
In the POC phase, the goal was to dissipate the object's energy smoothly, reducing and possibly eliminating the bouncing effect, which is a typical problem when considering stiff impacts. To this purpose, the robot's POC behavior was learned from human demonstrations by catching free-falling objects with unknown mass, and the learned behaviors, in terms of position and stiffness, were adapted to the task.

The whole procedure was tested through a series of comparative experiments. Firstly, we isolated the problem along a single dimension, where we showed how a catching task that would normally trigger the robot's safety brakes was successful with the proposed approach. We evaluated the results based on multiple metrics, and the best trade-off was found. We then studied the method's robustness by increasing the dropping height and eventually generalizing to multiple Cartesian axes. Several comparisons between the best methods and their alternatives with the addition of RMM were included. 

One limitation involves the event-driven perception system, since the camera has to be placed normally with respect to the trajectory plane of the flying object, due to the lack of in-depth perception.
These inaccuracies might lead to errors in the estimation of the catching time, resulting in a failed catch. These issues can be mitigated by
employing either stereo-camera or more accurate motion capture systems, however they remain an open issue in the perception field.

Regarding the human demonstration process, we asked the subject to use both arms to catch the free-falling object following the setup in~\cite{humanCatching2021} and calculated the mean value to estimate the human arm endpoint stiffness, which is eventually used in the experiments. 
Furthermore, the mass of the ball compared to the robot's maximum payload was proportional to the relationship between the weight of the box and the human subject's weight. With this setup, the robot achieved a good trade-off among the metrics in the VM-VIC scenario, in which the variable control gains of the VIC were scaled and filtered from HVS.
Once the weight of the ball was increased considerably, breaking the proportional relationship, we could no longer ensure similar results using the current learned HVS and the same releasing height, which is a limitation of the LfD part.
According to previous neuroscience research~\citep{lacquaniti1989role,zago2009visuo}, it is clear that humans actively change arm stiffness when intercepting a free-falling object with different contact momentum. However, the mechanism of transferring human stiffness to robots, in dynamic catching tasks, should be further explored and considered as a future research direction.

We then decided to maintain the orientation of the EE flat, to optimally isolate the catching forces, which are the pivotal point of our study, trying to avoid the additional dynamic components of self-induced EE forces and variations in the catching EE angles between multiple trials. Besides, we wanted to maintain a comparison between the single-axis (object dropping) and the generic multi-axis scenarios.
In the future, we will focus in detail on the multi-axis scenario, providing non-flat EE reference orientations, based on the measured object's velocity direction. 
Therefore, the same direction will be used for both planning and DIM maximization,
which is expected to further improve the beneficial characteristics of RMM achieved in the present (more conservative) work, potentially leading to smaller internal actuation torques. To achieve this, please note that no theoretical alteration is necessary, rather the estimated catching direction of the object (instead of the vertical one) is provided as a reference in 
\eqref{eq:general_QP_formulation} for the PRC planner.

Overall, there are multiple limitations related to this setup. 
For instance, regarding the controller, HQP-based schemes are known to suffer from discontinuities when transitioning between different SoT or constraints. While this was not necessary for this work, it is possible to exploit the HQP-based approach that we have previously developed in \cite{tassi_multibot} to ensure smooth transitions, enabling the possibility of switching to a different SoT after the impact is detected.
Besides, we are considering a specific nonprehensile catching task with a custom-made EE, which cannot cover all possible catching applications (Sec.~\ref{sec:intro}). Hence further generalization is necessary for real-life applicability. However, we believe that by focusing on the interaction forces, this work could provide a different insight for applications that involve high impacts and forces' exchange.

Future works will tackle the identification of the catching height 
based on DIM and other additional parameters,
considered not only at the control level but also at the planning stage.
Besides, non-horizontal EE orientations at the catching instant require further testing in a three-dimensional scenario, evaluating the effects of not considering a multi-dimensional impact model and neglecting state-jumps on the joint torques.

This will in turn enable a further extension, necessary to assess the generic nature of the proposed framework, which is to shift from a basket-like to a planar EE. This would imply a considerable increase in terms of complexity, since the catching success would be heavily affected by the precision in perception, trajectory estimation and controller's accuracy.
However, this could open to a wider variety of real-world applications, with better adaptability across multiple fields.

In conclusion, catching tasks are characterized by several complex aspects (pre- and post-planning, perception and trajectory estimation, stiffness regulation, impact-dynamics and induced joint torques, energy dissipation and object's bouncing) that need to be accurately addressed.
Firstly, an in-depth analysis is necessary to delve into each aspect, which is, with respect to optimal planning and impact analysis, where the present work finds place. Only after a distinct analysis of all the aforementioned critical aspects, their mutual effects can be studied, finally outlining a definitive overarching framework, capable of effectively achieving multi-purpose catching in most practical applications.

\begin{loa}
\begin{sortedlist}
\vspace{-1cm}
  \sortitem{\textbf{PRC}: Pre-Catching}
  \sortitem{\textbf{POC}: Post-Catching}
  \sortitem{\textbf{VM}: Velocity Matching}
  \sortitem{\textbf{HQP}: Hierarchical Quadratic Programming}
  \sortitem{\textbf{GMM}: Gaussian Mixture Model}
  \sortitem{\textbf{GMR}: Gaussian Mixture Regression}
  \sortitem{\textbf{QP}: Quadratic Programming}
  \sortitem{\textbf{SoT}: Stack of Tasks}
  \sortitem{\textbf{EE}: End-Effector}
  \sortitem{\textbf{RMM}: Reflected Mass Minimization}
  \sortitem{\textbf{VIC}: Variable Impedance Controller}
  \sortitem{\textbf{LfD}: Learning from human Demonstrations}
  \sortitem{\textbf{HVS}: Human arm physical Variable Stiffness}
  \sortitem{\textbf{CLIK}:Closed-loop Inverse Kinematics}
  \sortitem{\textbf{DIM}: Dynamic Impact Measure}
  \sortitem{\textbf{ADIM}: Average Dynamic Impact Measure}
  \sortitem{\textbf{LOI}: Lift Off Index}
  \sortitem{\textbf{DoF}: Degree of Freedom}
  \sortitem{\textbf{DRI}: Damping Ratio Index}
  \sortitem{\textbf{BTI}: Bouncing Time Index}
  \sortitem{\textbf{VME}: Velocity Matching Error}
  \sortitem{\textbf{FP}: Fixed-position}
  \sortitem{\textbf{IC}: Impedance Controller}
  \sortitem{\textbf{SIC}: Switching Impedance Controller}
  \sortitem{\textbf{IMA-Catcher}: IMpact-Aware Non-Prehensile Catching Framework}
  \sortitem{\textbf{RMS}: Root Mean Square}
  \sortitem{\textbf{PUCK}: Parallel surface and
Convolution Kernel}
\sortitem{\textbf{EROS}: Exponentially Reduced Ordinal Surface}

\end{sortedlist}
\end{loa}

\begin{funding}
This work was supported by the European Research Council's (ERC) starting grant Ergo-Lean (GA 850932).
\end{funding} 

\begin{dci}
The authors declare that they have no known competing financial interests or personal relationships that could have appeared to influence the work reported in this paper.
\end{dci}

\bibliographystyle{SageH}
\bibliography{biblio}

\newpage

\vspace{11pt}

\vfill

\end{document}